%% file: main.tex
\documentclass[runningheads]{llncs}

\usepackage{eccv}

\usepackage{eccvabbrv}

\usepackage{graphicx}
\usepackage{booktabs}

\usepackage[accsupp]{axessibility}  

\usepackage{multirow}
\usepackage{multicol}
\usepackage{wrapfig}
\usepackage{pifont}
\newcommand{\cmark}{\ding{51}}%
\newcommand{\xmark}{\ding{55}}%

\usepackage{dashrule}
\usepackage{import}
\usepackage{standalone}

\definecolor{blue_ppt}{rgb}{0.1914, 0.3359375, 0.9414}
\definecolor{ceruleanblue}{rgb}{0.16, 0.32, 0.75}

\usepackage{dsfont}

\newcommand{\SubItem}[1]{
    {\setlength\itemindent{15pt} \item[-] #1}
}

\usepackage{hyperref}

\usepackage{orcidlink}

\begin{document}

\title{Introducing Routing Functions to Vision-Language Parameter-Efficient Fine-Tuning with Low-Rank Bottlenecks} 

\titlerunning{Routing Functions For VL PEFT}

\author{Tingyu Qu\inst{1}\orcidlink{0000-0002-0656-5745} \and
Tinne Tuytelaars\inst{2}\orcidlink{0000-0003-3307-9723} \and
Marie-Francine Moens\inst{1}\orcidlink{0000-0002-3732-9323}}

\authorrunning{T.~Qu et al.}

\institute{Department of Computer Science, KU Leuven \and
Department of Electrical Engineering, KU Leuven \\
\email{\{tingyu.qu; tinne.tuytelaars; sien.moens\}@kuleuven.be}}

\maketitle

\begin{abstract}
  Mainstream parameter-efficient fine-tuning (PEFT) methods, such as LoRA or Adapter, project a model's hidden states to a lower dimension, allowing pre-trained models to adapt to new data through this low-rank bottleneck. 
  However, PEFT tasks involving multiple modalities, like vision-language (VL) tasks, require not only adaptation to new data but also learning the relationship between different modalities.
  Targeting at VL PEFT tasks, we propose a family of operations, called routing functions, to enhance VL alignment in the low-rank bottlenecks.
  These feature routing functions adopt linear operations and do not introduce new trainable parameters.
  In-depth analyses are conducted to study their behavior.
  In various VL PEFT settings,
  the routing functions significantly improve performance of the original PEFT methods, achieving over 20\% improvement on VQAv2 ($\text{RoBERTa}_{\text{large}}$+ViT-L/16) and 30\% on COCO Captioning (GPT2-medium+ViT-L/16).
  Also when fine-tuning a pre-trained multimodal model such as CLIP-BART, we observe smaller but consistent improvements across a range of VL PEFT tasks.
  Our code is available at \url{https://github.com/tingyu215/Routing_VLPEFT}.
  
\end{abstract}

\section{Introduction}
\label{sec:intro}

When fine-tuning pre-trained foundation models for downstream tasks, studies show that a relatively high accuracy can be achieved by acting on a much smaller dimension compared to the original dimension of the pre-trained model's feature space.
The minimum dimension necessary to approximate the optimization problem of the downstream task is termed the intrinsic dimension~\cite{li2018measuring, aghajanyan-etal-2021-intrinsic}.

\begin{wrapfigure}[26]{r}{0.5\textwidth}
  \begin{center}
  \includegraphics[width=0.5\columnwidth]{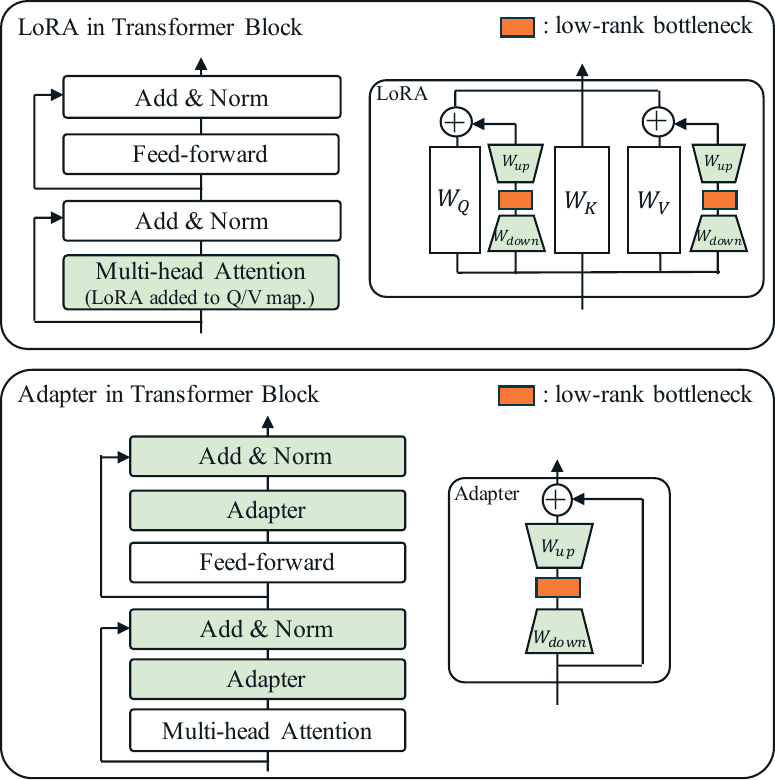}
  \end{center}
  \caption{Illustration of LoRA and Adapter used in a Transformer block. Only the green modules are updated in PEFT. We identify the low-rank bottleneck with the orange rectangle. The Adapters are sequentially added to the Transformer block.\protect \footnotemark}
  \label{fig:lora_Adapter}
\end{wrapfigure}

The existence of such intrinsic dimension explains the success of parameter-efficient fine-tuning (PEFT) methods with low-rank bottlenecks for model fine-tuning.
Two of the most widely applied techniques are Adapter~\cite{pmlr-v97-houlsby19a} and LoRA~\cite{hu2022lora}. 
We demonstrate the use of Adapter and LoRA within a Transformer block in~\cref{fig:lora_Adapter}. 
Despite being added at different positions within the Transformer block, Adapter and LoRA exhibit some key characteristics in their architectural design. Specifically, they first map the hidden states $x_H$ to a lower dimension $r$ through a down-projection mapping $W_{down}$. The down-projected states are then projected back to the original dimension $d$ via an up-projection mapping $W_{up}$.
In the PEFT context, only the parameters of the colored modules in~\cref{fig:lora_Adapter} are updated, while the remaining parameters are unchanged. Since the intermediate dimension $r$ is much smaller than the original dimension $d$, these methods significantly reduce the number of parameters that need to be updated.

In uni-modal tasks, where $x_H$ contains information from a single modality, $W_{down}$ effectively compresses $x_H$ into a feature space of lower dimension. This compression usually suffices provided that $r$ is larger than the intrinsic dimension of the downstream task~\cite{hu2022lora}.
However, PEFT methods with low-rank bottlenecks, such as Adapter and LoRA, are currently not designed for down-stream tasks involving another modality, such as vision-language (VL) tasks. Even for a pre-trained multimodal foundation model with $x_H$ containing information from both modalities, conceptually, directly compressing $x_H$ with $W_{down}$ poses challenges, as it is unclear how to effectively balance the two modalities and to enforce alignment between the modalities with this simple linear mapping. 
Therefore, in this paper we investigate whether it is possible to better route the features through the low-rank bottleneck with some extra guidance.
An intuitive approach is standard cross-attention. However, it introduces additional parameters through query/key/value projections. Moreover, softmax-based operations not only increase time complexity but are limited to dealing with features of length $>$ 1. This is incompatible with the PEFT setting, where the goal is to minimize the number of trainable parameters and computational complexity, while maintaining versatility across different types of tasks (including those that use only the [CLS] features).
In this work, we introduce feature routing functions in VL PEFT with low-rank bottlenecks. They rely on linear operations and do not introduce extra trainable parameters.

In summary, our contributions include:
\begin{itemize}
    \item We identify the problem of feature learning in the 
    low-rank bottlenecks of standard PEFT methods considering VL downstream tasks.
    \item We introduce a family of parameterized operations in the low-rank bottleneck, termed routing functions, that improve the performance of VL tasks.
    \item We conduct extensive experiments on different architectures and tasks, and provide 
    insights on using routing functions in VL PEFT tasks.
\end{itemize}

\section{Related Works}


\noindent \textbf{Adapter-based PEFT}
To reduce the tuning costs of pre-trained language models (LMs), the Adapter layer was first introduced in~\cite{pmlr-v97-houlsby19a}.
It first down-projects the model's hidden state, then up-projects it back to its original dimension, incorporating a nonlinear function after the down-projection.  Freezing the pre-trained weights of the LM and tuning them with sequentially added Adapter layers achieves performance comparable to full fine-tuning.
Since then, various Adapter variants have been proposed to advance the field of PEFT. 
The focus of developing Adapter variants has been on improving efficiency~\cite{ruckle-etal-2021-Adapterdrop, mahabadi2021compacter, lei2023conditional} and exploring alternative architectures~\cite{lin-etal-2020-exploring, karimi-mahabadi-etal-2021-parameter, he2022towards}.
Despite being surpassed by its variants on various benchmarks, the original sequential Adapter is still widely used in practice and offers stable results across tasks.

\noindent \textbf{LoRA-based PEFT}
LoRA~\cite{hu2022lora} adopts the same "down-projection then up-projection" approach as the Adapter, with no nonlinear function in the low-rank bottleneck.
LoRA is added to the query and value mappings in the attention modules of the pre-trained LM.
Important follow-up works include dynamic rank adjustment~\cite{valipour-etal-2023-dylora, zhang2023adaptive}
and quantization adaptation~\cite{NEURIPS2023_qlora, xu2024qalora}.
Moreover, VeRA~\cite{kopiczko2024vera} and NOLA~\cite{koohpayegani2024nola} further minimize the number of parameters by leveraging frozen random matrices.
LoRA-based PEFT methods have become standard for tuning large language models and large multimodal foundation models.

\noindent \textbf{Other PEFT methods}
Beyond Adapter and LoRA, prompt-based tuning methods~\cite{lester-etal-2021-power, li-liang-2021-prefix, jia2022vpt}
have shown strong performance for tuning both language and vision models.
LST~\cite{sung2022lst} learns a compact side network during tuning.
BitFit~\cite{ben-zaken-etal-2022-bitfit} only adjusts the bias terms during fine-tuning, and achieves comparable performance as other PEFT methods.
Methods~\cite{mao-etal-2022-unipelt, zhou-etal-2024-autopeft, he2022towards, wang-etal-2022-adamix} also explore combining PEFT methods to achieve superior performance.

\noindent \textbf{Vision-Language PEFT}
VL-Adapter~\cite{Sung_2022_CVPR} is the pioneering work in the field of Vision-Language (VL) PEFT. It empirically studies the use of different PEFT methods on four different VL tasks in a multitask learning setting.
VL-Adapter uses CLIP-BART, a similar model architecture as VL-T5~\cite{2020t5}. 
CLIP-BART uses a projection layer to a connect CLIP~\cite{pmlr-v139-radford21a} vision encoder to BART~\cite{lewis-etal-2020-bart} or T5~\cite{2020t5}. 
CLIP-BART is tuned using a combination of prompt tuning and other PEFT methods.
We conduct extensive experiments using CLIP-BART and VL-Adapter as presented in~\cref{sec:clip_bart}.
As an extension to VL-Adapter, VL-PET~\cite{Hu_2023_ICCV} combines an Adapter with a rather complicated granularity-controlled mechanism.
I-Tuning~\cite{ituning2023} explores alternative model architectures 
for VL PEFT tasks.
and proposes a parallel Adapter with cross-attention to integrate visual information into a pre-trained LM.
Since our work is the first that introduces routing functions into VL PEFT,  we limit the comparison of our method to VL-Adapter and I-Tuning, leaving the exploration of more complicated PEFT methods like VL-PET to future work.

\section{Methodology}

\begin{figure}[t]
  \centering
  \includegraphics[width=\linewidth]{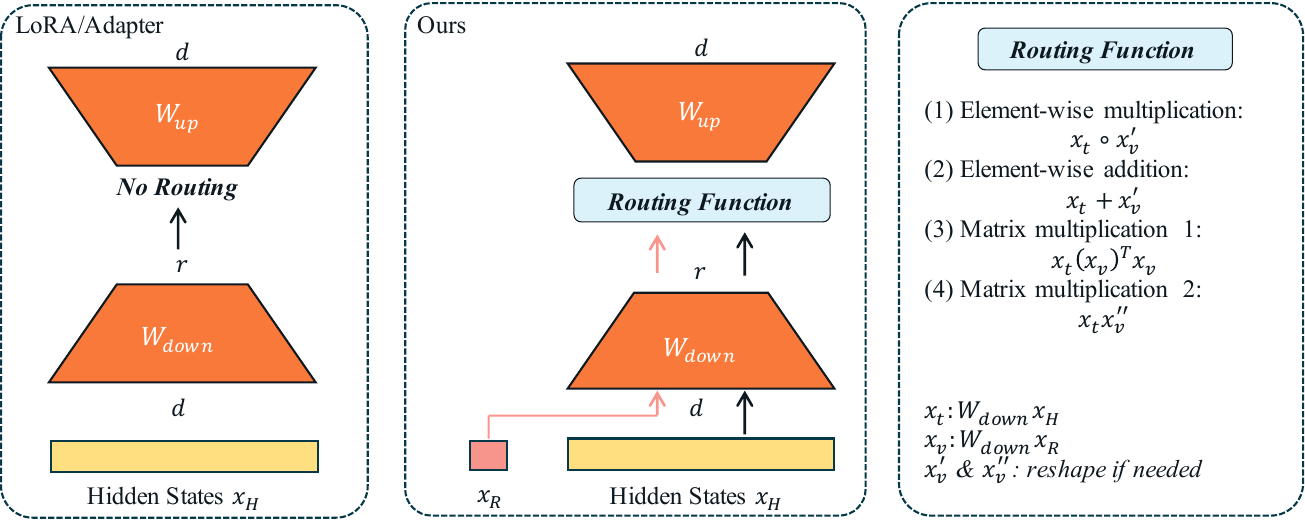}
  \caption{Illustration of our method. \textit{Left}:  Conventional PEFT methods with low-rank bottleneck first map the hidden states $x_H$ from a high dimension $d$ to a lower dimension $r$ via down-projection map $W_{down}$. Then, 
  it maps the hidden states back to the original dimension $d$ via 
  $W_{up}$. 
  \textit{Middle}: Our method utilizes the same architecture, but 
  updates the features 
  via a routing function in the low-rank bottleneck. Specifically in a VL task, for the features $x_R$ that we want to align to, we use $W_{down}$ to down-project them as $W_{down}x_R$.
  Then the routing function routes $W_{down}x_R$ and $W_{down}x_H$ in the low-rank bottleneck.
  \textit{Right}: Different routing functions. In Adapter, routing functions are added before nonlinear activation functions.
  }
  \label{fig:method}
\end{figure}

As illustrated in~\cref{fig:method} (Left), standard PEFT modules with low-rank bottleneck, \ie LoRA/Adapter, learn a down-projection mapping $W_{down}$ followed by an up-projection mapping $W_{up}$,
with no routing in between.
These modules are added in each Transformer block of the pre-trained foundation model (\cref{fig:lora_Adapter}).
As presented in~\cref{fig:method} (Middle), our method adds a routing function in the low-rank bottleneck. This routing function is designed to align two sets of features, 
in particular, the extra visual features $x_R$ to be integrated in the VL tasks and the hidden states $x_H$ of each Transformer block of the foundation model.

The design of a routing function follows two 'parameter-efficient' principles:
\begin{itemize}
    \item A linear operation is used to reduce computation overload.
    \item No extra trainable parameters are introduced.
\end{itemize}

In this paper, we explore four types of routing functions.
Given $x_H$ and $x_R$\footnote{Assuming $x_R$ to have the same dimension $d$ as $x_H$. See~\cref{sec:gpt_roberta} and~\cref{sec:clip_bart} for details.}, after the $W_{down}$ mapping, we obtain $x_t=W_{down}x_H$ and $x_v=W_{down}x_R$.
Here $x_t$ is of shape $(L_t, r)$ and $x_v$ is of shape $(L_v, r)$, where $L_t$/$L_v$ and $r$ are the feature length (\ie the number of tokens) and the bottleneck dimension or rank, respectively.\footnote{Here we omit the batch dimension for simplicity.}
In practice, we always use $L_v=1$, \ie we use a single token / global feature vector to represent the visual content. In case of multiple tokens, we use average pooling to reduce it to a single feature vector.\footnote{Only when comparing to cross-attention in~\cref{sec:vit_results}, we do not use pooled features.}
The proposed routing functions operate as:
\begin{itemize}
    \item Element-wise multiplication: $x_t\circ x'_v$, where $x'_v$ denotes $x_v.reshape\_as(L_t, r)$\footnote{\eg $x_v.reshape\_as(L_t, r)$: $x'_v=\mathds{1}^\mathsf{T} x_v$, where $\mathds{1}$ is a matrix of ones of shape $(L_v, L_t)$}
    \item Element-wise addition: $x_t+ x'_v$, where $x'_v$ denotes $x_v.reshape\_as(L_t, r)$ 
    \item Matrix multiplication 1: $x_t(x_v)^\mathsf{T}x_v$
    \item Matrix multiplication 2: $x_tx''_v$, where $x''_v$ denotes $x_v.reshape\_as(r, r)$
\end{itemize}

Intuitively, $x_t \circ x'_v$, reinforces or attenuates individual elements of $x_t$, using the corresponding element of $x_v$ as scaling factor.
 The second routing function, $x_t + x'_v$ replaces $x_t$ with the sum of $x'_v$ and $x_t$, thus moving $x_t$ closer to $x_v$. 
$x_t(x_v)^\mathsf{T}x_v$ projects $x_t$ onto $x_v$, \ie only keeping the component of $x_t$ that is aligned with $x_v$.
Finally, $x_t x''_v$ replaces $x_t$ with a rescaled version of $x_v$, with a different scale factor for each row given by the sum over all elements of the corresponding row of $x_t$. 
In all these cases, it should be kept in mind that both $x_v$ and $x_t$ are derived from $x_H$ and $x_R$ via the down-projection $W_{down}$. Parameters of $W_{down}$ can be learned so as to select particular dimensions of the original $x_H$ and $x_R$, on which to apply the above operations before up-projecting them again with $W_{up}$.

The features $x_R$ contain information from the modalities to which we aim to align $x_H$.
The choice of $x_R$ can be different for different architectures, for instance, [CLS] features from a ViT in case of an encoder-only or decoder-only LM.
We further explain the motivation of the choices of $x_R$ for different types of architectures in~\cref{sec:gpt_roberta_design} and \cref{sec:clip_bart_design}.

\section{Experiments with En(De)coder-Only Language Models}
\label{sec:gpt_roberta}

To explore the value of routing functions, in this section, we start from encoder-only and decoder-only language models (LMs)
and incorporate visual information from pre-trained vision encoders for vision-language (VL) parameter-efficient fine-tuning (PEFT) tasks.
The section begins with an overview of our experimental design and implementation details, and concludes with a discussion on the comprehensive experimental results obtained.

\subsection{Experimental Design}
\label{sec:gpt_roberta_design}

Here, we study VL PEFT tasks
in their cleanest form. 
Specifically, (1) No additional mapping networks are adopted, so the feature routing functions form the only module that contributes to the final performance; (2) Only the parameters of the PFET modules in LMs are updated, where we learn the VL alignment; (3) No multimodal tasks (\eg contrastive language-image pre-training in CLIP~\cite{pmlr-v139-radford21a}) are included in the pre-training phases of the 
backbone models.
We only learn the multimodal embedding space during VL PEFT.
These principles guarantee that VL alignment originates strictly from the PEFT modules.

\begin{figure}[t]
  \centering
  \includegraphics[width=\linewidth]{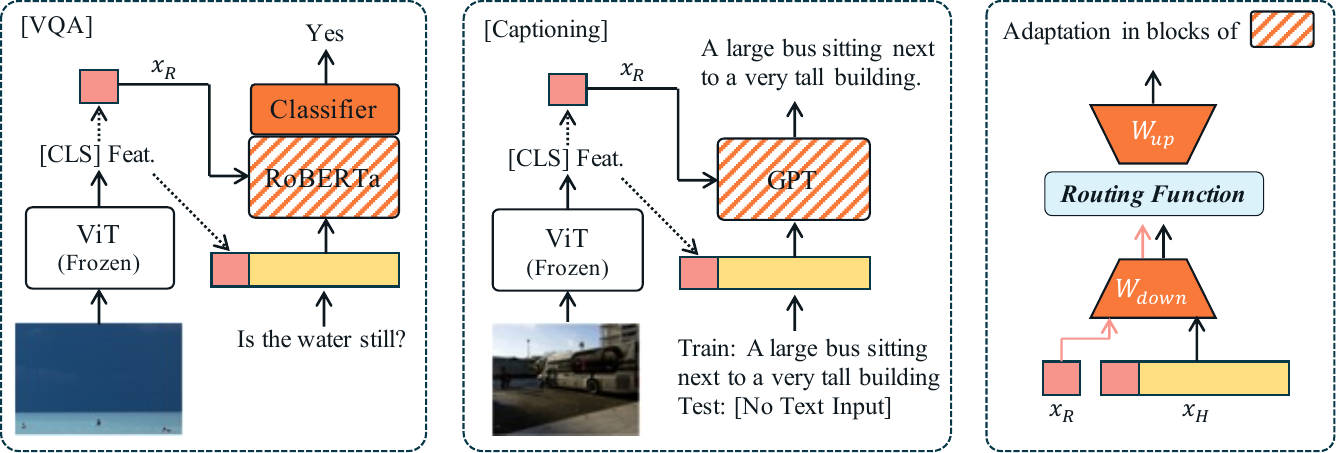}
  \caption{Pipeline for VQA and Image Captioning. Only the added classifier (in VQA) and the PEFT modules of the RoBERTa encoder and GPT decoder are tuned. 
  Note that here $x_H$ are the text inputs with the visual [CLS] features prepended.
  }
  \label{fig:enc_dec_model}
\end{figure}

To comprehensively study the role of routing functions in VL tasks,
we consider both discriminative and generative VL tasks.
We select visual question answering (VQA) and Image Captioning, two of the most widely studied VL tasks, as the representative task for discriminative VL tasks and generative VL tasks, respectively.

We present the pipelines for VQA and Image Captioning in~\cref{fig:enc_dec_model}.
We use ViT~\cite{dosovitskiy2021an} as our vision backbone. As for the LM backbone, we use the encoder-only RoBERTa~\cite{roberta19} and the decoder-only GPT2~\cite{radford2019language} for VQA and Image Captioning, respectively.\footnote{
In~\cref{sec:appendix_results_vit},
we report results obtained with larger backbones, \eg $\text{RoBERTa}_{\text{large}}$.} 
The [CLS] feature from the ViT is prepended to the text inputs
and acts as $x_R$, enabling the learning of VL alignment through PEFT, with or without the integration of our routing functions.
Prepending visual features to text inputs is actually the predominant way of learning VL alignment in PEFT~\cite{Sung_2022_CVPR, Hu_2023_ICCV} without routing functions.
The ViT and LMs adopt the same dimensions for their feature spaces, so  we do not need additional mappings to match the dimensions.

\subsection{Implementation Details}
\label{sec:vit_implementation}

We adopt ViT-B/16 pre-trained on ImageNet-21k as our vision backbone, which we keep frozen during training.
In LoRA, we set scaling factor $\alpha=r$ in all setups.
For VQAv2, we use $\text{RoBERTa}_{\text{base}}$ as our LM backbone,
with a two-layer multi-layer perceptron (MLP) classifier on top following~\cite{Sung_2022_CVPR}.
We train our model using the AdamW optimizer~\cite{loshchilov2018decoupled} with learning rate=3e-4 and batch size of 512 for 20 epochs, where we warm up the first 5\% of the steps.
For the captioning task, COCO Cap., we use GPT2-base as our LM backbone. 
We train our model for 20 epochs using the AdamW optimizer with batch size of 40 and learning rate=3e-5, where we warm up the first 5000 steps.
For more details, see~\cref{sec:appendix_details}.

\subsection{Results}

\label{sec:vit_results}

\noindent \textbf{Overall performance on VQA}
We first evaluate our routing functions on the discriminative task VQAv2 with $\text{RoBERTa}_{\text{base}}$ and ViT-B/16,
see~\cref{tab:roberta_vqa}.

\begin{wraptable}[13]{r}{0.5\linewidth}
  \centering  
  \caption{Accuracy and improvement over the baseline (in blue) obtained by different types of routing functions on VQAv2. The best results and second best results are in \textbf{bold} and \underline{underlined}, respectively.} 
    \resizebox{\linewidth}{!}{
    \begin{tabular}{lllll}
        \hline
        Routing & \multicolumn{2}{c}{$r=64$} & \multicolumn{2}{c}{$r=128$} \\
        Functions & LoRA & Adapter & LoRA & Adapter \\
        \hline
        None & 44.15 & 44.16 & 44.45 & 44.28 \\
        
        $x_t\circ x'_v$ & $53.51_{\color{blue_ppt}{+9.36}}$ & $52.78_{\color{blue_ppt}{+8.62}}$ & $\mathbf{53.86_{\color{blue_ppt}{+10.41}}}$ & $53.01_{\color{blue_ppt}{+8.73}}$ \\
        
        $x_t+x'_v$ & $52.60_{\color{blue_ppt}{+8.45}}$ & $53.94_{\color{blue_ppt}{+9.78}}$ &  $52.88_{\color{blue_ppt}{+8.43}}$ & $53.95_{\color{blue_ppt}{+9.67}}$ \\

        $x_t(x_v)^\mathsf{T}x_v$ & $\underline{53.88_{\color{blue_ppt}{+9.73}}}$ & $\underline{54.48_{\color{blue_ppt}{+10.32}}}$ & $\underline{53.09_{\color{blue_ppt}{+8.64}}}$ & $\mathbf{55.06_{\color{blue_ppt}{+10.78}}}$ \\

        $x_tx''_v$ & $\mathbf{54.21_{\color{blue_ppt}{+10.06}}}$ & $\mathbf{54.96_{\color{blue_ppt}{+10.80}}}$ &  $51.88_{\color{blue_ppt}{+7.43}}$ & $\underline{\color{blue_ppt}{54.38_{+10.10}}}$ \\
        
        \hline
    \end{tabular}
    }
    \label{tab:roberta_vqa} 
\end{wraptable}

Without the addition of routing functions, models with LoRA and Adapter reach a similar level of performance.
All four types of routing functions significantly improve the performance of the original PEFT method in terms of accuracy of the answer.
The dimension of the bottleneck does not have a strong impact on the final performance. Sometimes we can even obtain better performance with lower dimensions. 
It is a sign that the intrinsic dimension of the VL task is not high, just as observed in uni-modal tasks.
Moreover, we observe that the routing functions involving matrix multiplications (\ie $x_t(x_v)^\mathsf{T}x_v$ \& $x_tx''_v$) perform quite well in most scenarios for both LoRA and Adapter, while the routing functions involving element-wise operations (\ie $x_t\circ x'_v$ \& $x_t + x'_v$) are just slightly behind.

\begin{table}[!htbp]
    \centering
    \begin{minipage}{0.46\textwidth}
        \centering
        \caption{Results in terms of accuracy with different routing functions on VQAv2 sorted by answer type. The "all" column reports overall accuracy.}
        \resizebox{\linewidth}{!}{
        \begin{tabular}{lllll | llll}
        \hline
        Routing & \multicolumn{4}{c}{LoRA, $r=128$} & \multicolumn{4}{c}{Adapter, $r=128$} \\
        Functions & all &  yes/no & number & other & all & yes/no & number & other \\
        \hline
        None & 44.45 & 68.67 & 31.29 & 29.20 & 44.28 &  68.38 & 31.19 & 29.11  \\

        $x_t\circ x'_v$ & $\mathbf{53.86}$ & $\mathbf{72.29}$ & $\underline{35.36}$ & $\mathbf{44.48}$ & 53.01 & 71.72 & 33.34 & 43.72 \\
        
        $x_t+x'_v$ & 52.88 & $\underline{72.03}$ & 34.20 & 43.00 & 53.95 & 72.43 & $\mathbf{35.38}$ & 44.56 \\
        
        $x_t(x_v)^\mathsf{T}x_v$ & $\underline{53.09}$ & 70.63 & $\mathbf{35.40}$ & $\underline{44.19}$ & $\mathbf{55.06}$ & $\mathbf{74.09}$ & 33.59 & $\mathbf{45.99}$ \\        
        
        $x_tx''_v$ & 51.88 & 69.92 & 33.29 & 42.82 & $\underline{54.38}$ & $\underline{73.00}$ & $\underline{33.94}$ & $\underline{45.35}$ \\
        
        \hline
    \end{tabular}
    }
    \label{tab:roberta_vqa_anstype}
    \end{minipage}\hfill
    \begin{minipage}{0.53\textwidth}
        \centering
        \caption{Results in terms of accuracy with  different routing functions on VQAv2 sorted by question type. 
        The "all" column reports overall accuracy.
        }
    \resizebox{\linewidth}{!}{
    \begin{tabular}{llllll | lllll}
        \hline
        Routing & \multicolumn{5}{c}{LoRA, $r=128$} & \multicolumn{5}{c}{Adapter, $r=128$} \\
        Functions & all& where & color & time & person & all & where & color & time & person \\
        \hline
        None & 44.45 & 22.64  & 40.00  & 14.29  & 45.17 & 44.28 & 22.93  & 40.15  & 13.70  & 44.94 \\

        $x_t\circ x'_v$ & $\mathbf{53.86}$ & 35.70  & $\underline{51.69}$  & $\mathbf{23.07}$ &  $\underline{62.74}$ & 53.01 & 35.59  & 52.63  & 20.73  & 62.28 \\
        
        $x_t+x'_v$ & 52.88 & $\underline{36.12}$ & 49.33  & 21.22 & $\mathbf{62.95}$ & 53.95 & $\mathbf{37.08}$  & 52.08  & 21.66  & $\underline{63.76}$ \\
        
        $x_t(x_v)^\mathsf{T}x_v$ & $\underline{53.09}$ & $\mathbf{36.33}$  &  $\mathbf{52.86}$  &  22.29  &  61.74 & $\mathbf{55.06}$ & $\underline{36.92}$ & $\mathbf{53.66}$  & $\mathbf{24.44}$  & $\mathbf{63.85}$  \\
        
        $x_tx''_v$ & 51.88 & 34.19 & 51.60  & $\underline{22.63}$  & 61.14 & $\underline{54.38}$ & 35.64 & $\underline{52.77}$  & $\underline{21.85}$  & 62.50 \\
        
        \hline
    \end{tabular}
    }
    \label{tab:roberta_vqa_questype}
    \end{minipage}
\end{table}

To dive deeper into the behaviors of different types of routing functions, we evaluate the performance on different answer types as presented in~\cref{tab:roberta_vqa_anstype}.
The largest improvements are obtained for the "other" category for all four routing functions.
It can be seen that different routing functions excel for different answer types. 
For example, even when the two element-wise operations reach lower overall accuracy, they lead to higher or comparable accuracy on number-type answers. 
As for the routing functions involving matrix multiplications, better performance on other-type answers is obtained.
For instance, with relatively lower overall accuracy reached in LoRA, matrix multiplications reach comparable performance on other-type answers.
No clear pattern can be observed for yes/no types of answers, which could be explained by the large variety in content of these questions.

\cref{tab:roberta_vqa_questype} shows our method's performance across various question types.
Four distinct question types test the method's reasoning capabilities concerning fine-grained visual information: spatial information ("where", \eg Where is the surfboard?), color information ("color", \eg What color is the bike?), time information ("time", \eg What time is on the clock?), and person information ("person", \eg What is the man holding?).\footnote{The comprehensive list of question types is included 
in~\cref{sec:appendix_details}.
}
Adding routing functions significantly improves the performance of LoRA and Adapter.
Notably, different routing functions seem to capture different aspects of the visual information.
$x_t + x'_v$ shows great performance while reasoning on spatial information (where) and person information (person). It is assumed that the addition operator reinforces feature values of content both present in the question and image, which might be beneficial for spatial and person questions.
Similarly, $x_t x''_v$ performs exceptionally well for time information (time).
With $x_t \circ x'_v$, we obtain relatively better performance on the color information (color).
$x_t (x_v)^\mathsf{T} x_v$ presents consistent strong abilities while aligning spatial information (where) and color information (color).

In~\cref{sec:appendix_results_vit}, 
we show an additional ablation study on replacing visual features ($x_R$) with random noise or all-ones features.
We also compare inference time of LoRA/Adapter with LoRA/Adapter + routing functions, where we observe adding routing functions to PEFT methods does not reduce efficiency.

\begin{table}[t]
  \centering 
    \caption{Results of different types of routing functions on COCO Cap. The Avg. score is the averaged result from BLEU-4, METEOR, ROUGE-L, CIDEr, and SPICE. The best and second best results are in \textbf{bold} and \underline{underlined}, respectively.} 
    \resizebox{\linewidth}{!}{
    \begin{tabular}{lllllll | llllll}
        \hline
        Routing & \multicolumn{6}{c}{LoRA, $r$=64} & \multicolumn{6}{c}{LoRA, $r$=128} \\
        Functions & BLEU-4 & METEOR & ROUGE-L & CIDEr & SPICE & Avg. & BLEU-4 & METEOR & ROUGE-L & CIDEr & SPICE & Avg. \\
        \hline
        None & 18.3 & 17.4 & 36.8 & 55.8 & 11.8 & 28.02 & 21.0 & 20.9 & 43.4 & 70.3 & 14.7 & 34.06 \\

        $x_t\circ x'_v$ & $\mathbf{26.1_{\color{blue_ppt}{+7.8}}}$ & $\mathbf{23.7_{\color{blue_ppt}{+6.3}}}$ & $\mathbf{48.6_{\color{blue_ppt}{+11.8}}}$ & $\mathbf{88.7_{\color{blue_ppt}{+32.9}}}$ & $\mathbf{17.3_{\color{blue_ppt}{+5.5}}}$ & $\mathbf{40.88_{\color{blue_ppt}{+12.86}}}$ & $\underline{25.4_{\color{blue_ppt}{+4.4}}}$ & $\mathbf{23.7_{\color{blue_ppt}{+2.8}}}$ & $\mathbf{48.2_{\color{blue_ppt}{+4.8}}}$ & $\mathbf{87.9_{\color{blue_ppt}{+17.6}}}$ & $\mathbf{17.7_{\color{blue_ppt}{+3.0}}}$ & $\mathbf{40.58_{\color{blue_ppt}{+6.52}}}$  \\

        $x_t+x'_v$ & $22.2_{\color{blue_ppt}{+3.9}}$ & $20.3_{\color{blue_ppt}{+2.9}}$ & $41.0_{\color{blue_ppt}{+4.2}}$ & $73.5_{\color{blue_ppt}{+17.7}}$ & $14.8_{\color{blue_ppt}{+3.0}}$ & $34.36_{\color{blue_ppt}{+6.34}}$ & $\underline{25.4_{\color{blue_ppt}{+4.4}}}$ & $\underline{23.1_{\color{blue_ppt}{+2.2}}}$ & $\underline{47.1_{\color{blue_ppt}{+4.7}}}$ & $\underline{87.1_{\color{blue_ppt}{+16.8}}}$ & $17.0_{\color{blue_ppt}{+2.3}}$ & $\underline{39.94_{\color{blue_ppt}{+5.88}}}$ \\

        $x_t(x_v)^\mathsf{T}x_v$ & $\underline{24.8_{\color{blue_ppt}{+6.5}}}$ & $\underline{22.6_{\color{blue_ppt}{+5.2}}}$ & $\underline{45.4_{\color{blue_ppt}{+8.6}}}$ & $\underline{84.9_{+29.1}}$ & $\underline{16.8_{\color{blue_ppt}{+5.0}}}$ & $\underline{38.90_{\color{blue_ppt}{+10.88}}}$ & $24.9_{\color{blue_ppt}{+3.9}}$ & $22.7_{\color{blue_ppt}{+1.8}}$ & $45.4_{\color{blue_ppt}{+2.0}}$ & $83.8_{\color{blue_ppt}{+13.5}}$ & $16.9_{\color{blue_ppt}{+2.2}}$ & $38.74_{\color{blue_ppt}{+4.68}}$ \\
        
        $x_tx''_v$ & $23.9_{\color{blue_ppt}{+5.6}}$ & $21.9_{\color{blue_ppt}{+4.5}}$ & $43.9_{\color{blue_ppt}{+7.1}}$ & $80.5_{\color{blue_ppt}{+14.7}}$ & $16.2_{\color{blue_ppt}{+4.4}}$ & $37.28_{\color{blue_ppt}{+9.26}}$ & $\mathbf{25.5_{\color{blue_ppt}{+4.5}}}$ & $22.9_{\color{blue_ppt}{+2.0}}$ & $46.0_{\color{blue_ppt}{+2.6}}$ & $85.2_{\color{blue_ppt}{+14.9}}$ & $\underline{17.2_{\color{blue_ppt}{+2.5}}}$ & $39.36_{\color{blue_ppt}{+5.30}}$ \\
        \hline
        Routing & \multicolumn{6}{c}{Adapter, $r$=64} & \multicolumn{6}{c}{Adapter, $r$=128} \\
        Functions & BLEU-4 & METEOR & ROUGE-L & CIDEr & SPICE & Avg. & BLEU-4 & METEOR & ROUGE-L & CIDEr & SPICE & Avg. \\
        \hline
        None & 15.9 & 18.5 & 37.0 & 61.6 & 14.1 & 29.42 & 16.1 & 18.4 & 36.2 & 61.5 & 13.9 & 29.22 \\

        $x_t\circ x'_v$  & $\underline{24.6_{\color{blue_ppt}{+8.7}}}$ & $\underline{23.1_{\color{blue_ppt}{+4.6}}}$ & $46.4_{\color{blue_ppt}{+9.4}}$ & $84.5_{\color{blue_ppt}{+22.9}}$ & $\underline{17.2_{\color{blue_ppt}{+3.1}}}$ & $39.16_{\color{blue_ppt}{+9.74}}$ & $23.6_{\color{blue_ppt}{+7.5}}$ & $\underline{22.6_{\color{blue_ppt}{+4.2}}}$ & $\underline{44.8_{\color{blue_ppt}{+8.6}}}$ & $\underline{82.8_{\color{blue_ppt}{+21.3}}}$ & $\underline{17.1_{\color{blue_ppt}{+3.2}}}$ & $\underline{38.18_{\color{blue_ppt}{+8.96}}}$  \\

        $x_t+x'_v$ & $21.0_{\color{blue_ppt}{+5.1}}$ & $21.5_{\color{blue_ppt}{+3.0}}$ & $42.4_{\color{blue_ppt}{+5.4}}$ & $75.0_{\color{blue_ppt}{+14.4}}$ & $16.0_{\color{blue_ppt}{+1.9}}$ & $35.18_{\color{blue_ppt}{+5.76}}$ & $19.3_{\color{blue_ppt}{+3.2}}$ & $20.8_{\color{blue_ppt}{+2.4}}$ & $40.9_{\color{blue_ppt}{+4.7}}$ & $75.2_{\color{blue_ppt}{+13.8}}$ & $16.2_{\color{blue_ppt}{+2.3}}$ & $34.48_{\color{blue_ppt}{+5.26}}$  \\

        $x_t(x_v)^\mathsf{T}x_v$ & $\underline{26.1_{+10.2}}$ & $\mathbf{23.2_{\color{blue_ppt}{+4.7}}}$ & $\mathbf{46.9_{\color{blue_ppt}{+9.9}}}$ & $\underline{85.4_{\color{blue_ppt}{+23.8}}}$ & $\mathbf{17.3_{\color{blue_ppt}{+3.2}}}$ & $\underline{39.78_{\color{blue_ppt}{+10.36}}}$ & $\underline{24.1_{\color{blue_ppt}{+8.0}}}$ & $22.4_{\color{blue_ppt}{+4.0}}$ & $44.2_{\color{blue_ppt}{+8.0}}$ & $82.7_{\color{blue_ppt}{+21.2}}$ & $16.9_{\color{blue_ppt}{+3.0}}$ & $38.06_{\color{blue_ppt}{+8.84}}$  \\
        
        $x_tx''_v$ & $\mathbf{26.6_{\color{blue_ppt}{+10.7}}}$ & $23.0_{\color{blue_ppt}{+4.5}}$ & $\underline{46.8_{\color{blue_ppt}{+9.8}}}$ & $\mathbf{85.8_{\color{blue_ppt}{+24.2}}}$ & $\underline{17.2_{\color{blue_ppt}{+3.1}}}$ & $\mathbf{39.88_{\color{blue_ppt}{+10.46}}}$ & $\mathbf{26.0_{\color{blue_ppt}{+9.9}}}$ & $\mathbf{23.5_{\color{blue_ppt}{+5.1}}}$ & $\mathbf{46.8_{\color{blue_ppt}{+10.6}}}$ & $\mathbf{86.3_{\color{blue_ppt}{+24.8}}}$ & $\mathbf{17.8_{\color{blue_ppt}{+3.9}}}$ & $\mathbf{40.08_{\color{blue_ppt}{+10.86}}}$ \\
        \hline
    \end{tabular}
    }
    \label{tab:gpt2_cap} 
\end{table}

\noindent \textbf{Overall performance on image captioning}
Evaluation of image captioning relies on the overall assessment of the generated captions.
We report multiple text generation metrics and their average scores in~\cref{tab:gpt2_cap}. The addition of routing functions again leads to significant performance gains for all metrics in all scenarios.
As was the case with VQAv2, different patterns are observed.
Changing the intermediate dimension $r$ from 128 to 64 seems to have a larger impact on the model performance with conventional LoRA, where without the routing functions the CIDEr score drops from 70.3 to 55.8.
In contrast, models augmented with routing functions remain robust and performant across setups, despite the changes in intermediate dimension. 
$x_t\circ x'_v$ and $x_t x''_v$ perform well, possibly due to their alignment power. 
We also report the performance of Adapter with larger backbones, namely GPT2-medium and ViT-L/16, 
in~\cref{sec:appendix_results_vit}.
Consistent and significant improvements in performance are observed, showing the efficacy of our routing functions on the task.
\begin{wraptable}[14]{r}{0.5\linewidth}
  \centering 
  \caption{
  Comparing routing functions with cross-attention
  on COCO Cap. $\dag$: parallel Adapter as used in~\cite{ituning2023}. Separate Map.: separate mapping for query/key/value (in cross-attn.) or features $x_R$/$x_H$ (in $x_t(x_v)^\mathsf{T}x_v$). Here $r=128$. Avg. indicates the same average scores as in~\cref{tab:gpt2_cap}. Full evaluation 
  in~\cref{sec:appendix_results_vit}.
  } 
    \resizebox{\linewidth}{!}{
    \begin{tabular}{lclllll}
        \hline
        PEFT & Separate Map. & Alignment & Param. & BLEU-4 & CIDEr & Avg. \\
        \hline
        LoRA & \cmark & Cross-attn. & 4.786M & 28.7 & 92.2 & 43.02 \\
        LoRA & \cmark & $x_t (x_v)^\mathsf{T} x_v$ & \textbf{3.932M} & $30.7_{\color{blue_ppt}{+2.0}}$ & $99.4_{\color{blue_ppt}{+7.2}}$ & $45.68_{\color{blue_ppt}{+2.66}}$ \\
        LoRA & \xmark & $x_t (x_v)^\mathsf{T} x_v$ & \textbf{4.746M} & $30.0_{\color{blue_ppt}{+1.3}}$ & $99.0_{\color{blue_ppt}{+6.8}}$ & $45.38_{\color{blue_ppt}{+2.36}}$ \\
        \hline
         $\text{Adapter}^\dag$ & \cmark & Cross-attn. & 4.732M & 30.7 & 99.8 & 45.70 \\
        $\text{Adapter}^\dag$ & \xmark  & $x_t (x_v)^\mathsf{T} x_v$ & \textbf{1.830M} & $30.8_{\color{blue_ppt}{+0.1}}$ & $98.8_{-1.0}$ & $45.48_{-0.22}$ \\
        \hline
    \end{tabular}
    }
    \label{tab:gpt_crossattn} 
\end{wraptable}

\noindent \textbf{Comparison to cross-attention}
Here we compare the results of the integration of routing function $x_t (x_v)^\mathsf{T} x_v$ with these of integrating cross-attention between the textual and the visual representation in image captioning. Indeed, in principle, a cross-attention function can also be considered as a kind of routing function with nonlinear operations.

I-Tuning~\cite{ituning2023} implements an Adapter with cross-attention for image captioning. 
To have a fair comparison, we follow the settings of I-Tuning for both LoRA and Adapter. 
Specifically, (1) instead of using the visual features, we prepend the "<s>" token to the input text. (2) We use the last hidden state from ViT as key and value for cross-attention and as $x_R$ for routing functions. This change is needed, since the [CLS] feature is of length 1, which does not work for cross-attention due to the Softmax function.

In the case of LoRA, we also implement a variant with separate $W_{down}$ mappings for $x_R$ and $x_H$.
This brings LoRA with routing function closer to LoRA with cross-attention.\footnote{ 
\cref{sec:appendix_details} 
shows detailed architectures and explains why routing functions outperform cross-attention.
We down-project to $r/4$ and $r$ for LoRA w/ \& w/o separate mapping, respectively, to have the same level of parameter count.
}
Without separate mapping refers to the setting used in previous experiments, where $x_R$ and $x_H$ share the same $W_{down}$.
As shown in~\cref{tab:gpt_crossattn}, despite of having less parameters, LoRA+$x_t (x_v)^\mathsf{T} x_v$ (w/ separate mapping) achieves significantly better performance than LoRA with cross-attention. 
The setup of LoRA+$x_t (x_v)^\mathsf{T} x_v$ (w/o separate mapping) outperforms LoRA with cross-attention by a large margin using a similar amount of parameters.

For Adapter, the cross-attention module has the same intermediate dimension as the $W_{down}$ in Adapter with routing function.
Naturally, the separate query/key/value mappings make that the Adapter with 
cross-attention has more than double the number of parameters compared to the Adapter equipped with the routing function ($x_t (x_v)^\mathsf{T} x_v$). 
Both achieve comparable performance (\cref{tab:gpt_crossattn}).

To conclude, \cref{tab:gpt_crossattn} shows that integrating 
$x_t (x_v)^\mathsf{T} x_v$ in LORA and Adapter yields better performance given the same amount of parameters, or is similar in performance while using substantially less parameters. 
For VL tasks like image captioning, integrating a routing function to guide feature representations in low-rank bottlenecks has superior abilities compared to cross-attention.

\begin{wraptable}[10]{r}{0.5\linewidth}
  \centering 
  \caption{Experiments on adding nonlinear activation functions in routing functions. We use $r=64$ here.} 
    \resizebox{\linewidth}{!}{
    \begin{tabular}{llcccc}
        \hline
        PEFT  & Routing &  VQA & \multicolumn{3}{c}{COCO Cap.} \\
        Method & Functions & Accuracy & BLEU-4 & CIDEr & Avg. \\
        \hline
        LoRA & $x_t (x_v)^\mathsf{T} x_v$ & 53.88 & 24.8 & 84.9 & 38.90 \\
        LoRA & $ReLU(x_t) ReLU(x_v)^\mathsf{T} x_v$ & 53.45 & 25.3 & 86.1 & 39.84 \\
        Adapter & $x_t (x_v)^\mathsf{T} x_v$ & 54.48 & 26.1 & 85.4 & 39.78 \\
        Adapter & $ReLU(x_t) ReLU(x_v)^\mathsf{T} x_v$ & 54.63 & 25.0 & 84.7 & 39.02 \\
        \hline
    \end{tabular}
    }
    \label{tab:relu} 
\end{wraptable}

\noindent \textbf{Incorporating nonlinear activation functions into routing functions}
We have proposed linear routing functions for efficiency reasons, but they do not need to be exclusively linear.  
Taking inspirations from the linear attention variants~\cite{pmlr-v119-katharopoulos20a, choromanski2021rethinking, zhen2022cosformer}, where the linear operation is often equipped with nonlinear activation functions to ensure non-negativity for similarity scores between query and key, we explore whether a similar operation affects the representation power of the routing functions in the low-rank bottleneck.
For routing function $x_t(x_v)^\mathsf{T}x_v$, we use the nonlinear activation function $ReLU(\cdot)$ as in previous work~\cite{zhen2022cosformer}, and transform it to $ReLU(x_t)ReLU(x_v)^\mathsf{T}x_v$.
The results are presented in~\cref{tab:relu}.
With or without $ReLU(\cdot)$, our routing function achieves similar performance on VQA and image captioning.

\begin{figure}[!htbp]
  \centering
  \includegraphics[height=2.55cm]{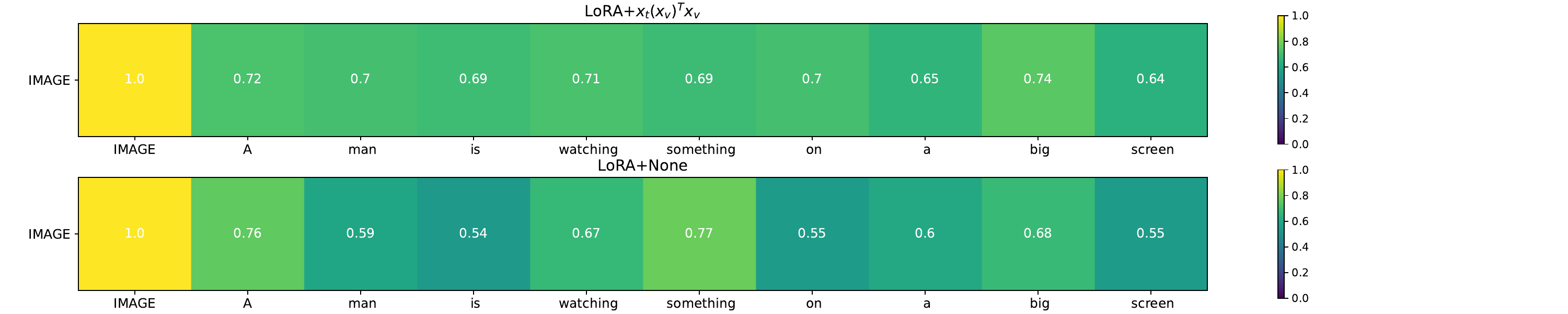}
  \caption{Example of average attention weights from the last layer of GPT2. We use the final checkpoint trained on COCO Cap. with $r=64$. IMAGE: visual [CLS] feature.}
  \label{fig:example}
\end{figure}

\noindent \textbf{Qualitative Analysis}
We first qualitatively compare the VL alignment pattern of LoRA-tuned GPT2 with or without a routing function on image captioning. An example is presented in~\cref{fig:example}.
With the addition of routing function $x_t (x_v)^\mathsf{T} x_v$, the model pays more attention to the nouns (\eg man, a big screen) and the verbs (\eg watching), which are essential in learning the VL alignment. 
At the same time, it pays less attention to meaningless words like "something". This is in line with the finding that the routing function stresses components of $x_t$ that are aligned in the image.

\begin{figure}[t]
  \centering
  \includegraphics[width=0.9\linewidth]{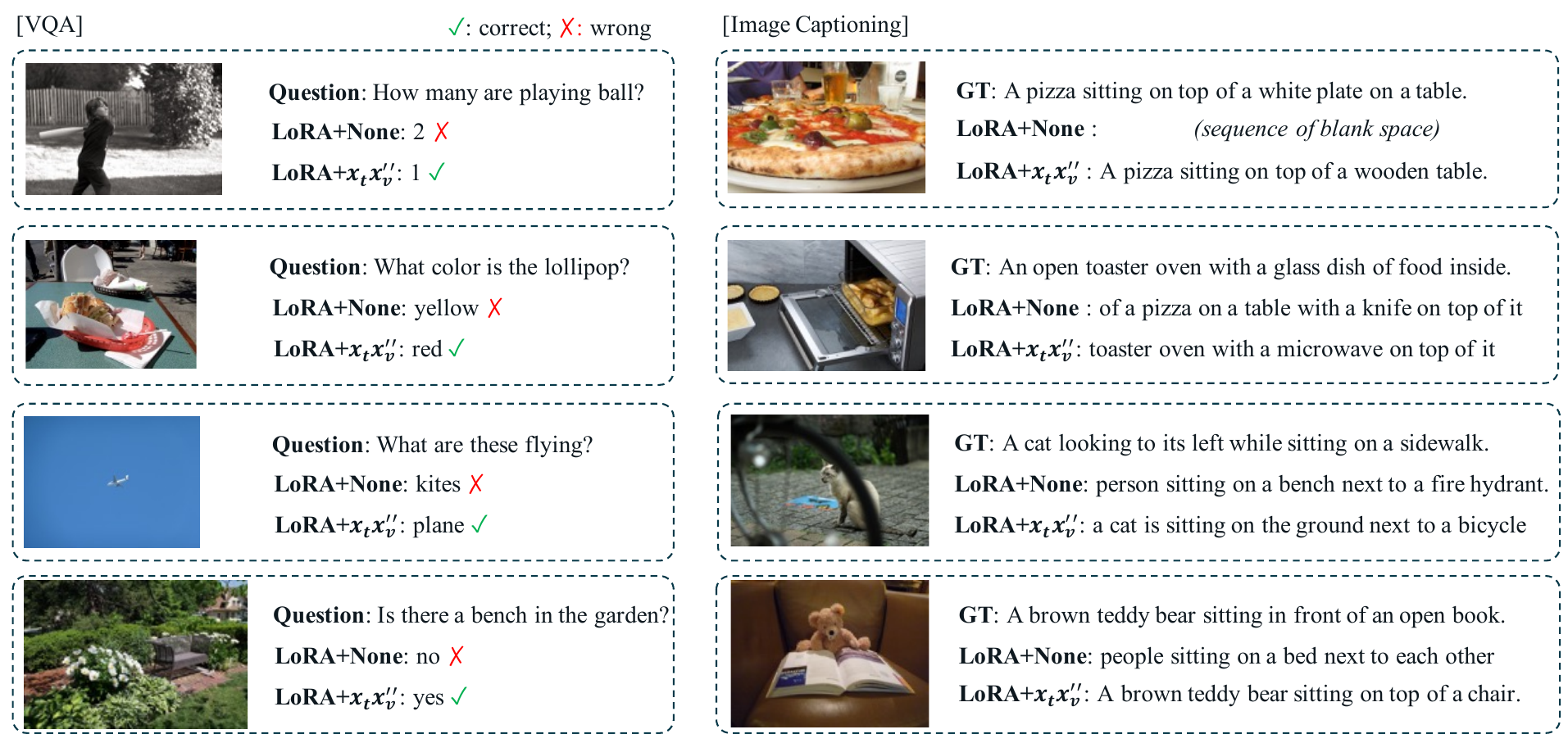}
  \caption{Qualitative examples from VQA and Image Captioning tasks. 
  We present cases where the model trained with conventional LoRA fails and results obtained from a model trained with LoRA with a routing function.
  We use $r=64$.}
  \label{fig:gen_example}
\end{figure}

Next we qualitatively examine some examples from the inference phase for VQA and Image Captioning.
We present examples in~\cref{fig:gen_example} where conventional LoRA fails.
For VQA, previous experiments in~\cref{tab:roberta_vqa_questype} indicate that our routing functions work particularly well on color related questions.
As also shown in~\cref{fig:gen_example}, conventional LoRA fails to align the color red to the lollipop.
The model tuned with conventional LoRA also fails to correctly answer other types of questions as presented in the figure.
Without routing functions, captions generated from the LM tuned with conventional LoRA often make mistakes in comprehending the visual information.
Occasionally, we observe a sequence of blank spaces as presented in the first example using conventional LoRA.

\section{Experiments with Encoder-Decoder Language Models}
\label{sec:clip_bart}

In this section, we study the VL PEFT tasks in a more complicated setting, where we consider an encoder-decoder LM with an additional mapping network dedicated for the vision encoder. The model is trained in a multi-task learning setting, where four different VL tasks are jointly learned, as in~\cite{cho2021vlt5, Sung_2022_CVPR}. 
Despite the increased complexity of optimizing the network in this scenario, we show that integrating routing functions into the low-rank bottlenecks yields performance enhancements. 
Moreover, consistent patterns are observed for different scenarios, which shed lights on future research to improve VL PEFT tasks.

\subsection{Experimental Design}
\label{sec:clip_bart_design}

\begin{wrapfigure}[24]{r}{0.5\textwidth}
  \begin{center}
  \includegraphics[width=0.5\columnwidth]{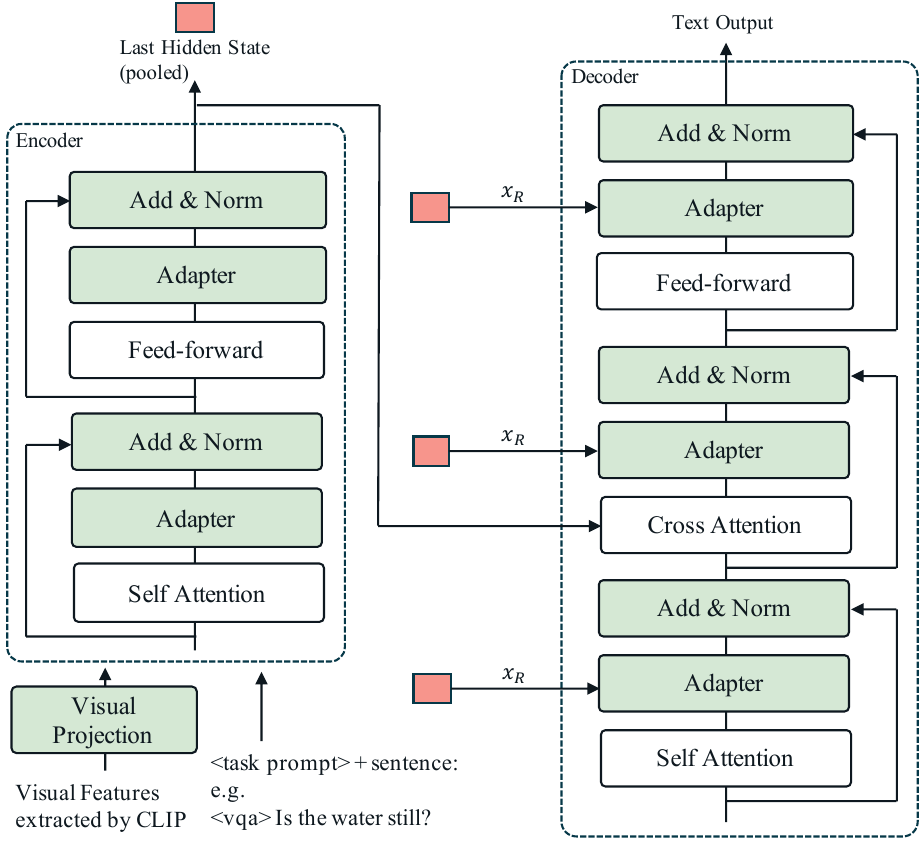}
  \end{center}
  \caption{Architecture of CLIP-BART with Adapter. The standard Adapter architecture is adopted (see~\cref{fig:lora_Adapter}). Only the modules in green are trainable. To apply routing functions, we use the pooled last encoder hidden state as feature $x_R$, which is added to the Adapters in the decoder.}
  \label{fig:clip_bart}
\end{wrapfigure}

Following~\cite{Sung_2022_CVPR}, we use the CLIP-BART architecture to study the effect of our routing functions on the VL PEFT tasks in a multitask learning setting.
CLIP-BART is a widely used architecture for VL tasks. We present the simplified model architecture of CLIP-BART with Adapters and the integrated routing functions in~\cref{fig:clip_bart}.
The original CLIP-BART uses a trainable visual projection module to map the visual features extracted by CLIP to the embedding space of BART~\cite{lewis-etal-2020-bart}.
CLIP-BART uses task-specific prompts as the textual input: For example, in~\cref{fig:clip_bart}, "<vqa>" is added as prompt to the text for VQA.
The mapped visual features are then prepended to the textual input. Together, they serve as the input to BART.
CLIP-BART is trained and evaluated in a multitask setting, that is, VQAv2~\cite{balanced_vqa_v2} and GQA~\cite{Hudson_2019_CVPR} for visual question answering, $\text{NLVR}^2$~\cite{suhr-etal-2019-corpus} using natural language for visual reasoning, and COCO Cap.~\cite{mscoco} for image captioning.
CLIP-BART unifies the four tasks into one text generation problem following~\cite{cho2021vlt5}, where four different task prompts distinguish the task types.
CLIP-BART can be considered as the representative of the current state-of-the-art on VL PEFT tasks.

To integrate our routing functions into the model, two types of inputs $x_R$ can be considered, namely the visual features after the visual projection, and the last encoder hidden state as presented in~\cref{fig:clip_bart}.
The visual features only contain visual information.
Because the last encoder hidden state contains both textual and visual information, it can only be added to the decoder.
We experiment with both LoRA and Adapter as the PEFT method for fine-tuning CLIP-BART.
In this multitask learning setting, one can use (1) a single LoRA/Adapter shared across tasks or (2) multiple LoRAs/Adapters without parameter sharing, where each task has its own LoRA/Adapter.
We start by exploring the impact of different routing functions on the single LoRA/Adapter setting.
Then, based on the observations obtained in the setting with single Adapter, we explore ways of using the routing functions in the multiple Adapters setting.

\subsection{Implementation Details}
\label{sec:clip_bart_implementation}
Following~\cite{Hu_2023_ICCV, Sung_2022_CVPR}, we use CLIP-ResNet101 as our vision encoder~\cite{pmlr-v139-radford21a}. 
We use $\text{BART}_{\text{base}}$ as our encoder-decoder LM as in~\cite{Sung_2022_CVPR}.
Following~\cite{Sung_2022_CVPR}, we use $r=96$ for all experiments.
The training pipeline in~\cite{Sung_2022_CVPR} is used for a fair comparison.
Specifically, (1) we use the same seed for all experiments.
(2) Due to limited resources, we do not perform an extensive hyper-parameter search as in~\cite{Sung_2022_CVPR}.
(3) We train the model with AdamW optimizer and linear decay for 20 epochs.
We warm-up the learning rate from 0 to the highest in the first 2 epochs.
(4) The batch size is set to 500. 
For more details, we refer to~\cref{sec:appendix_details}.

\subsection{Results}
\label{sec:clip_bart_results}

\noindent \textbf{Experiments with a single Adapter/LoRA}
We first report results of experiments on using single Adapter/LoRA shared across tasks in~\cref{tab:clip_bart_single}.
In this case, the Adapter/LoRA is trained with data from four different tasks, and naturally we can only use one type of routing function throughout the experiments.

\begin{table}[t]
  \centering 
    \caption{Experiments with different types of routing functions using CLIP-BART with single LoRA/Adapter.
    Following~\cite{Hu_2023_ICCV} ($\dag$) \protect \footnotemark, we report the average CIDEr score for COCO Cap., accuracy for the other three tasks.
    We also report the average accuracy from VQA, GQA and $\text{NLVR}^2$.
    We \textbf{bold} and \underline{underline} the best and second best results. 
    } 
    \resizebox{\linewidth}{!}{
    \begin{tabular}{l | llll | l | llll | l}
        \hline
        Routing & \multicolumn{5}{c}{Single LoRA} & \multicolumn{5}{c}{Single Adapter} \\
        Functions & VQA & GQA & $\text{NLVR}^2$ & Avg. & COCO Cap.  & VQA & GQA & $\text{NLVR}^2$ & Avg. & COCO Cap. \\
        \hline
        $\text{None}^\dag$ & 65.15 & 53.66 & 72.58 & 63.80 & 115.01  & 65.76 & 54.16 & 73.19 & 64.37 & 114.61  \\
        
        \hline

        $x_t\circ x'_v$ & $\mathbf{65.68}$ & $\mathbf{53.96}$ & 73.42 & $\mathbf{64.35}$ & 113.94 & $\mathbf{65.92}$ & $\mathbf{54.34}$ & $\underline{74.23}$ & $\mathbf{64.83}$ & 114.38 \\

        $x_t+x'_v$ & $\underline{65.14}$ & $\underline{53.73}$ & $\underline{73.51}$  & $\underline{64.13}$ & 114.96 & $\underline{65.89}$ & $\underline{54.18}$ & 73.90 & $\underline{64.66}$ & 114.39 \\

        $x_t(x_v)^\mathsf{T}x_v$ & 64.94 & 53.56 & $\mathbf{73.60}$ & 64.03 & $\underline{117.80}$ & 65.84 & 53.65 & $\mathbf{74.31}$ & 64.03 & $\underline{117.65}$ \\
        
        $x_tx''_v$ & 64.84 & 53.13 & 72.98 & 63.65 & $\mathbf{119.26}$ & 65.83 & 53.61 & 73.27 & 64.24 & $\mathbf{118.86}$  \\
        \hline
    \end{tabular}
    }
    \label{tab:clip_bart_single} 
\end{table}

\cref{tab:clip_bart_single} shows that the integration of the routing functions in the single LoRa and single Adapter architecture results in improved performance
in nearly all scenarios.
But just like the experiments with encoder-only and decoder-only LMs in~\cref{sec:vit_results}, different routing functions show different patterns.
The routing functions involving matrix multiplication, that is, $x_t (x_v)^\mathsf{T} x_v$ and $x_t x''_v$, perform exceptionally well in the COCO captioning task, while the other two routing functions involving element-wise operations, \ie $x_t \circ x'_v$ and $x_t+x'_v$, show more potential on QA tasks (VQA/GQA).
As for the $\text{NLVR}^2$ task, $x_t (x_v)^\mathsf{T} x_v$ and $x_t \circ x'_v$ present stronger abilities to learn VL relationships.
Note that achieving relatively low performance on one task does not necessarily mean that the corresponding routing function fails.
For example, $x_t \circ x'_v$ does not work well for captioning, but achieves great performance on the other three tasks. This somehow contrasts with the results obtained with encoder-only and decoder-only LMs in~\cref{sec:vit_results}, where
$x_t \circ x'_v$ leads to one of the best performance on COCO Cap.
The complexity of the multitask learning setup, where different loss functions are jointly optimized, might here be a confounding factor.

\footnotetext{\dag: 3 seeds.~\cref{sec:appendix_results_clip_bart} shows that our improvements are significant using paired t-test with $p=0.05$.}

\noindent \textbf{Experiments with multiple Adapters}
We further explore the behavior of the routing functions in models with multiple Adapters,
where we train a separate Adapter per task. 
As also observed in~\cite{Sung_2022_CVPR}, using multiple Adapters increases the complexity of the model, and yields worse performance compared to using a single Adapter.
As presented in~\cref{tab:clip_bart_multiple}, using the four routing functions with multiple Adapters leads to comparable or better performance as compared to multiple Adapters with no routing functions.
Moreover, in line with what we discovered in the experiments with models with a single Adapter, different routing functions shine on different types of tasks.

In this set-up,
we also experiment with combinations of routing functions.
The four combinations integrate the routing functions that work best for each individual task (\eg $x_t x'_v$ for COCO Cap.). The results in~\cref{tab:clip_bart_multiple} evidence that in this multitask setting combinations of routing functions 
yield an improved performance compared to using no routing functions. 
GQA requires more fine-grained image representation. For efficiency reason, we only adopt $x_v$ as the down-projection from the pooled last encoder hidden state. Nevertheless, comb1 can recover from this shortcoming. More detailed image representations as used in the cross-attention experiments of~\cref{tab:gpt_crossattn} might still enhance the results.

Overall, our research has focused on introducing routing functions in low-rank feature approximations when fine-tuning models for downstream (in our case VL) tasks based on pre-trained foundation models. As a pioneering work on the topic, for obtaining results we did not exhaustively search for all possible choices of hyperparameters. 
Instead, we have focused on revealing the problem of low-rank bottlenecks and on providing a successful solution. 
We leave an exhaustive grid search for optimal hyperparameters as future work.

\begin{table}[t]
  \centering 
    \caption{Experiments with different combinations of routing functions using CLIP-BART with multiple Adapters. 
    Following~\cite{Sung_2022_CVPR} ($\ddag$), we report CIDEr score for COCO Cap., accuracy for the other three tasks.
    We also report the average accuracy from VQA, GQA and $\text{NLVR}^2$.
    We \textbf{bold} and \underline{underline} the best and second best results.
    } 
    \resizebox{\linewidth}{!}{
    \begin{tabular}{l | c | c | c | c | llll | l}
        \hline
        Routing & \multicolumn{4}{c}{Routing Functions} & \multicolumn{5}{c}{Performance} \\
        Functions & VQA & GQA & $\text{NLVR}^2$ & COCO Cap.  & VQA & GQA & $\text{NLVR}^2$ & Avg. & COCO Cap.  \\
        \hline
        $\text{None}^\ddag$ & \xmark & \xmark & \xmark & \xmark & 65.4 & 54.0 & 69.8 & 63.07 & 114.3 \\
        
        \hline
        
        $x_t\circ x'_v$ &  \cmark & \cmark & \cmark & \cmark & 65.61 & 53.51 & 70.78 & $\underline{63.30}$ & 113.72 \\

        $x_t+x'_v$ &  \cmark & \cmark & \cmark & \cmark & 65.19 & $\underline{53.84}$ & $\underline{70.99}$ & 63.34 & 115.64 \\

        $x_t(x_v)^\mathsf{T}x_v$ & \cmark & \cmark & \cmark & \cmark & 64.93 & 52.11 & 70.25 & 62.43 & 116.20 \\
        
        $x_tx''_v$ &  \cmark & \cmark & \cmark & \cmark & 65.32 & 52.67 & 70.58 & 62.86 & $\underline{117.92}$ \\

        \hline

        comb1 & $x_t\circ x'_v$ & $x_t\circ x'_v$ & $x_t(x_v)^\mathsf{T}x_v$ & $x_tx''_v$ & $\mathbf{66.37}$ & $\mathbf{54.21}$ & 69.27 & 63.28 & 116.94  \\

        comb2 & $x_t+x'_v$ & $x_t+x'_v$ & $x_t(x_v)^Tx_v$ & $x_tx''_v$ & 65.14 & 53.37 & 70.88 & 63.13 & 117.58 \\

        comb3 & $x_t\circ x'_v$ & $x_t\circ x'_v$ & $x_t\circ x'_v$ & $x_tx''_v$ & $\underline{66.34}$ & 53.77 & $70.02$ & $\mathbf{63.38}$ & 117.73 \\

        comb4 & $x_t + x'_v$ & $x_t + x'_v$ & $x_t\circ x'_v$ & $x_tx''_v$ & 65.47 & 52.99 & $\mathbf{71.01}$ & 63.16 & $\mathbf{118.34}$  \\

        \hline
    \end{tabular}
    }
    \label{tab:clip_bart_multiple} 
\end{table}

\section{Conclusion}

We have introduced feature routing functions to help guide the feature learning in low-rank bottlenecks in architectures that are typical for parameter-efficient fine-tuning such as LoRA and Adapter. We have shown their validity in vision-language tasks, where we have focused on linear functions that do not require any extra parameters. We have demonstrated performance gains in single-task and multi-task settings in combination with a variety of foundation models.

Routing functions open many avenues for future research. Given the importance of foundation models that are trained in a self-supervised way on large data, there is the need to efficiently fine-tune them for downstream tasks. When computing low-rank approximations of the features, our work shows that the extra guidance of the routing functions is important. We see many possibilities in experimenting with other types of functions in other tasks. For instance, linear and other functions are also suited to model constraints that reflect external knowledge or expectations on the behavior of a model. Future work will show whether they are useful to guide the feature routing in low-rank bottlenecks.

\section*{Acknowledgements}
This work is funded by the Flanders AI Research Program, European Research Council Advanced Grant H2020-ERC-2017-ADG 788506, and the China Scholarship Council.

%
%
\bibliographystyle{splncs04}
\bibliography{main}

\clearpage

\input{supplementary}

\end{document}

%% file: supplementary.tex
\title{Introducing Routing Functions to Vision-Language Parameter-Efficient Fine-Tuning with Low-Rank Bottlenecks: Appendix} 

\titlerunning{Routing Functions for VL PEFT}

\author{Tingyu Qu\inst{1}\orcidlink{0000-0002-0656-5745} \and
Tinne Tuytelaars\inst{2}\orcidlink{0000-0003-3307-9723} \and
Marie-Francine Moens\inst{1}\orcidlink{0000-0002-3732-9323}}

\authorrunning{T.~Qu et al.}

\institute{Department of Computer Science, KU Leuven \and
Department of Electrical Engineering, KU Leuven \\
\email{\{tingyu.qu; tinne.tuytelaars; sien.moens\}@kuleuven.be}}

\maketitle

\appendix

\renewcommand{\thetable}{\Alph{section}\arabic{table}}
\renewcommand{\thefigure}{\Alph{section}\arabic{figure}}

\section{Additional Implementation Details}
\label{sec:appendix_details}

\noindent \textbf{List of VQA question types}

For experiments on VQAv2~\cite{balanced_vqa_v2} in~\cref{sec:vit_results}(\cref{tab:roberta_vqa_questype}),
we list the specific question types involved in "what", "color", "time" and "person" below:

\begin{itemize}
    \item \textit{where}: 
    \SubItem{where is the; where are the; what is on the; what is in the}
    \SubItem{Example: What is on the benches?}
    \item \textit{color}: 
    \SubItem{what color (what color is the;  what color is; what color are the); what is the color of the;}
    \SubItem{Example: What color is the bird?}
    \item \textit{time}: 
    \SubItem{what time}
    \SubItem{Example: What time does the clock read?}
    \item \textit{person}: 
    \SubItem{what is the man; what is the woman; is the man; is the woman; what is the person; is the person; is this person}
    \SubItem{Example: What is the woman pushing?}
\end{itemize}

\begin{figure}[!htbp]
  \centering
  \includegraphics[width=\linewidth]{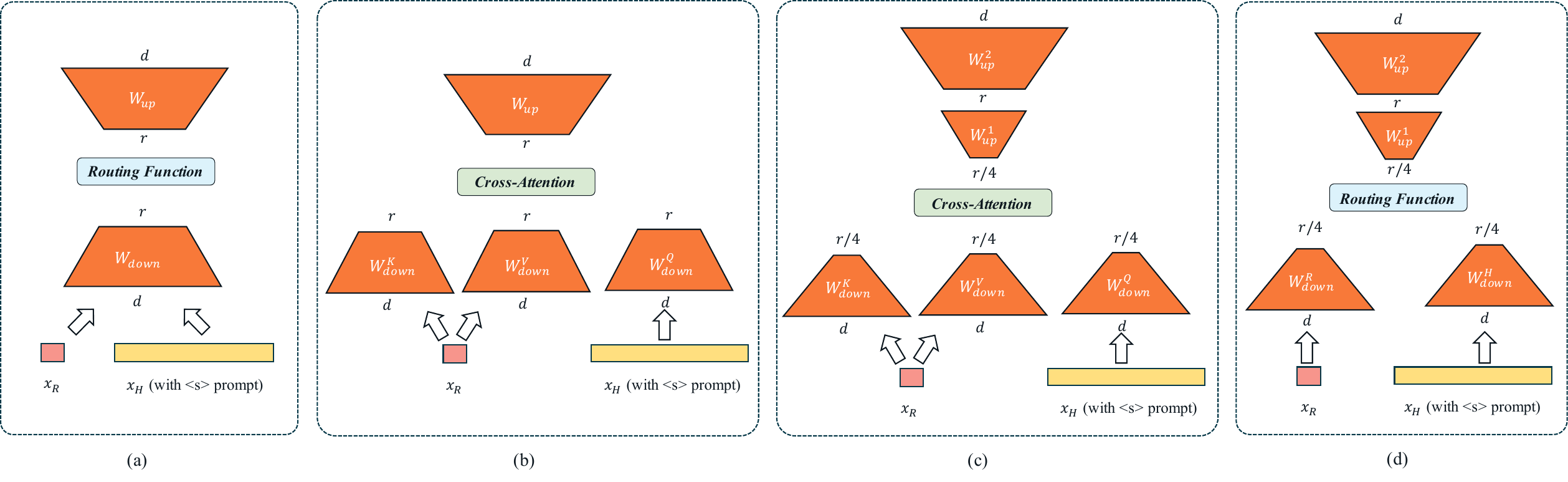}
  \caption{Illustration of architectures used in comparison to cross-attention. 
  (a) Standard way of integrating routing functions in PEFT methods. We implement this in Adapter/LoRA w/o separate mapping.
  (b) Cross-attention integrated in Adapter as introduced in~\cite{ituning2023}. We implement this in Adapter + cross-attention following~\cite{ituning2023}.
  (c) Cross-attention in PEFT, but with additional up-projection map which project the hidden states to $r$ first. We implement this in LoRA with cross-attention.
  (d) PEFT with routing funcitons using separate mapping for vision and language. We implement this in LoRA with separate mapping to examine how will separate mappings in cross-attention affect performance.}
  \label{fig:crossattn}
\end{figure}

\noindent \textbf{Architectures with separate mapping}
We present the illustration of the architectures used for experiments on comparing the routing functions to cross-attention in~\cref{fig:crossattn}.
The experiments can be found in~\cref{sec:vit_results} (\cref{tab:gpt_crossattn}).
We also implement a variant with separate $W_{down}$
mappings for $x_R$ and $x_H$.

To keep the number of parameters comparable to our standard setting, in LoRA + cross-attention as shown in~\cref{fig:crossattn} (c), we down-project to $r/4$. After cross-attention, we first up-project to rank $r$, which is the same as the standard setting. Finally we up project back to the original dimension $d$.
In that case, to have a fair comparison, in LoRA + routing functions (w/ separate mapping) as shown in~\cref{fig:crossattn} (d), we follow exactly the same mapping strategy, namely $d\rightarrow r/4\rightarrow r \rightarrow d$, which results in less trainable parameters as compared to our standard setting and the cross-attention in~\cref{fig:crossattn} (c).
For detailed parameter count, we refer to~\cref{tab:gpt_crossattn} in the main text.

\noindent \textbf{Additional implementation details for RoBERTa/GPT2 + ViT}
In addition to~\cref{sec:vit_implementation}, we report additional implementation details of RoBERTa/GPT2 + ViT in this part.
We use the Karpathy split of VQAv2~\cite{balanced_vqa_v2} and COCO Captioning~\cite{mscoco} (COCO Cap.) in our experiments.
For VQAv2, only the parameters of the classifier and the PEFT modules in RoBERTa are updated during training.
For COCO Cap., only the parameters of the PEFT modules in GPT2 are updated during training.
The maximum length for captions is set to 50.
After training, we evaluate the models using the final checkpoint for both tasks.

\noindent \textbf{Additional implementation details for CLIP-BART}
In addition to~\cref{sec:clip_bart_implementation}, we report additional implementation details of CLIP-BART in this part.
The same procedure for extracting visual features is adopted as in~\cite{Sung_2022_CVPR}.
Following~\cite{Sung_2022_CVPR}, we extract the 7$\times$7 grid features from the last convolutional layer of the vision encoder, which we down-project to 6$\times$6 use adaptive maximum-pooling.
The image resolution is set to 224$\times$224 following~\cite{Sung_2022_CVPR}.
For CLIP-BART experiments, we conduct our experiments based on the official implementation of VL-Adapter~\cite{Sung_2022_CVPR}, which searches for optimal learning rate among \{$1\times10^{-4}$, $3\times10^{-4}$, $1\times10^{-3}$\}. Limited to resources, we only search for optimal learning rate among \{$3\times10^{-4}$, $1\times10^{-3}$\}. The adopted learning rates for different experiments are presented in~\cref{tab:learning_rate}.

\begin{table}[t]
  \centering 
    \caption{Learning rates used for experiments with CLIP-BART.} 
    \resizebox{\linewidth}{!}{
    \begin{tabular}{lllllllll}
        \hline
        PEFT & $x_t\circ x'_v$ & $x_t + x'_v$ & $x_t(x_v)^\mathsf{T}x_v$ & $x_t x''_v$ & comb1 & comb2 & comb3 & comb4 \\
        \hline
        Single LoRA & $1\times10^{-3}$ & $1\times10^{-3}$ & $1\times10^{-3}$ & $1\times10^{-3}$ &\\
        Single Adapter & $1\times10^{-3}$ & $1\times10^{-3}$ & $1\times10^{-3}$ & $1\times10^{-3}$ & \\ 
        Multiple Adapter & $3\times10^{-4}$ & $3\times10^{-4}$ & $3\times10^{-4}$ & $3\times10^{-4}$ & $1\times10^{-3}$ & $3\times10^{-4}$ & $1\times10^{-3}$ & $3\times10^{-4}$ \\
        
        \hline
    \end{tabular}
    }
    \label{tab:learning_rate} 
\end{table}

\section{Additional Results with RoBERTa/GPT2 + ViT}
\label{sec:appendix_results_vit}

Supplementary to the experiments presented in~\cref{sec:vit_results}, in this section, we report additional experimental results on (1) replacing features $x_R$ on VQAv2; (2) comparing average inference time; (3) experiments with larger backbones; (4) complete evaluation of comparing routing functions to cross-attention and (5) pattern analysis of $W_{down}$/$W_{up}$ mappings. 

\begin{wraptable}[10]{r}{0.4\linewidth}
  \centering 
  \caption{Experiments on replacing $x_R$ on VQAv2.} 
    \resizebox{\linewidth}{!}{
    \begin{tabular}{llll}
        \hline
        Routing & \multirow{2}{*}{$x_R$} & \multicolumn{2}{c}{r=64} \\
        Functions & & LoRA & Adapter \\
        \hline
        None & None & 44.15 & 44.16 \\
        $x_t(x_v)^\mathsf{T}x_v$ & vis. [CLS] & $53.88_{+9.73}$ & $54.48_{+10.32}$ \\
        $x_t(x_v)^\mathsf{T}x_v$ & random noise & $43.80_{-0.35}$ & $43.48_{-0.68}$ \\
        $x_t(x_v)^\mathsf{T}x_v$ & ones & $45.59_{+1.44}$ & $24.42_{-19.74}$ \\ 
        
        \hline
    \end{tabular}
    }
    \label{tab:roberta_vqa_replase_vis} 
\end{wraptable}

\noindent \textbf{Are we learning the correct alignment?}
To answer this question, we replace the $x_R$ of choice (visual [CLS] features) with random noise or features with all ones.
Both types of features lead to meaningless VL alignment.
If we are learning the correct VL alignment via our routing functions in the low-rank bottleneck, changing the feature $x_R$ in this way should result in either no significant improvements or degradation in performance.
We present the results with $x_t(x_v)^\mathsf{T}x_v$ in~\cref{tab:roberta_vqa_replase_vis}.
Clearly we can see, random noise features or features with all ones lead to degradation in performance in most cases. The small improvement in performance obtained by aligning $x_t$ to features with all ones using LoRA with r=64 is far worse than the improvement made by using the original $x_R$ (visual [CLS] features).
Especially when we switch from LoRA to Adapter using the same features with all ones, we see significant degradation in performance (-19.74 in accuracy).
The experiments with replacing the features show that the significant improvements that our method made indeed come from learning a good alignment.

\begin{wraptable}[9]{r}{0.4\linewidth}
  \vspace{-\intextsep}
  \hspace*{-.75\columnsep}
  \centering 
  \caption{Average inference time (ms) per sample with $r=64$.} 
    \resizebox{\linewidth}{!}{
    \begin{tabular}{lllll}
        \hline
        Routing & \multicolumn{2}{c}{VQAv2} & \multicolumn{2}{c}{COCO Cap.} \\
        Functions &  LoRA & Adapter &  LoRA & Adapter \\
        \hline
        None & 0.2375 & 0.2539 & 17.7432 & 17.2814 \\
        $x_t\circ x'_v$ & 0.2533 & 0.2542 & 17.6533 & 18.3681 \\
        $x_t+ x'_v$ & 0.2527 & 0.2475 & 17.8562 & 18.8359 \\
        $x_t(x_v)^\mathsf{T}x_v$ & 0.2536 & 0.2543 & 18.2282 & 18.9063 \\
        $x_tx''_v$ & 0.2568 & 0.2584 & 18.4309 & 19.0058 \\

        \hline
    \end{tabular}
    }
    \label{tab:inference_time} 
\end{wraptable}

\noindent \textbf{Are we sacrificing the efficiency?}
By introducing the routing functions into the low-rank bottleneck of LoRA and Adapter, we are also interested in how it will affect the inference efficiency.
We compute the actual inference time (in ms) of our models on a RTX3090 GPU using the torch.cuda.event() api and report the results in~\cref{tab:inference_time}.
The addition of our routing functions does not bring significant increase in inference time, which shows their suitability for PEFT tasks.

\begin{wraptable}[9]{r}{0.5\linewidth}
  \vspace{-\intextsep}
  \hspace*{-.75\columnsep}
  \centering  
  \caption{Accuracy obtained by different types of routing functions with Adapter (r=64) on VQAv2. We vary the backbone sizes.} 
    \resizebox{\linewidth}{!}{
    \begin{tabular}{cc | ccccc}
        \hline
         \multicolumn{2}{c}{Backbone} & \multicolumn{5}{c}{Routing Functions}  \\
         Vision & Language & None & $x_t(x_v)^\mathsf{T}x_v$ & $x_t\circ x'_v$ & $x_t+x'_v$ & $x_tx''_v$  \\
        \hline
        ViT-B/16 & $\text{RoBERTa}_{\text{base}}$ &  44.16 & 54.48 & 52.78 & 53.94 & 54.96 \\
        ViT-L/16 & $\text{RoBERTa}_{\text{large}}$ & 44.82 & 55.43 & 54.83 & 52.06 & 55.92  \\
        
        \hline
    \end{tabular}
    }
    \label{tab:roberta_vqa_large} 
\end{wraptable}

\noindent \textbf{Experiments with larger backbones}
To evaluate how model scale affects the final performance of the routing functions, we conduct experiments with larger backbones.
Specifically, for VQAv2, we replace the vision backbone ViT-B/16 with ViT-L/16, and the language backbone $\text{RoBERTa}_{\text{base}}$ with $\text{RoBERTa}_{\text{large}}$.
It makes the dimension of the model hidden states increase from 768 to 1024. 
We report the results with Adapter (r=64) in~\cref{tab:roberta_vqa_large}.
The same conclusion still holds, that is, all four routing functions can significantly increase the performance of the vanilla Adapter.
Similarly, we conduct experiments using GPT2-medium and ViT-L/16 for COCO Cap., and report the results in~\cref{tab:gpt2_cap_all_adapter_large}.
As expected, the integration of routing functions bring significant improvements in performance.

\begin{table}[!htbp]
  \centering 
    \caption{Experiments with different types of routing functions with Adapter (r=64) on COCO Cap. We compare results obtained with different model sizes. The Avg. score for COCO Cap. is the averaged result from BLEU-4, CIDEr, METEOR, ROUGE-L and SPICE. The best results and the second best results are \textbf{bold} and \underline{underlined}, respectively.} 
    \resizebox{\linewidth}{!}{
    \begin{tabular}{lllllll | llllll}
        \hline
        Routing & \multicolumn{6}{c}{GPT2-base+ViT-B/16} & \multicolumn{6}{c}{GPT2-medium+ViT-L/16} \\
        Functions & BLEU-4 & METEOR & ROUGE-L & CIDEr & SPICE & Avg. & BLEU-4 & METEOR & ROUGE-L & CIDEr & SPICE & Avg. \\
        \hline
        None & 15.9 & 18.5 & 37.0 & 61.6 & 14.1 & 29.42 & 20.1 & 19.1 & 42.7 & 62.1 & 12.8 & 31.36 \\

        $x_t\circ x'_v$  & $24.6_{\color{blue_ppt}{+8.7}}$ & $\underline{23.1_{\color{blue_ppt}{+4.6}}}$ & $46.4_{\color{blue_ppt}{+9.4}}$ & $84.5_{\color{blue_ppt}{+22.9}}$ & $\underline{17.2_{\color{blue_ppt}{+3.1}}}$ & $39.16_{\color{blue_ppt}{+9.74}}$ & $22.1_{\color{blue_ppt}{+2.0}}$ & $21.7_{\color{blue_ppt}{+2.6}}$ & $42.8_{+0.1}$ & $75.4_{\color{blue_ppt}{+13.3}}$ & $16.5_{\color{blue_ppt}{+3.7}}$ & $35.70_{\color{blue_ppt}{+3.34}}$ \\

        $x_t+x'_v$ & $21.0_{\color{blue_ppt}{+5.1}}$ & $21.5_{\color{blue_ppt}{+3.0}}$ & $42.4_{\color{blue_ppt}{+5.4}}$ & $75.0_{\color{blue_ppt}{+14.4}}$ & $16.0_{\color{blue_ppt}{+1.9}}$ & $35.18_{\color{blue_ppt}{+5.76}}$ & $25.3_{\color{blue_ppt}{+5.2}}$ & $\underline{23.3_{\color{blue_ppt}{+4.2}}}$ & $\underline{47.6_{\color{blue_ppt}{+4.9}}}$ & $\underline{88.2_{\color{blue_ppt}{+26.1}}}$ & $\underline{17.5_{\color{blue_ppt}{+4.7}}}$ & $\underline{40.38_{\color{blue_ppt}{+9.02}}}$ \\

        $x_t(x_v)^\mathsf{T}x_v$ & $\underline{26.1_{\color{blue_ppt}{+10.2}}}$ & $\mathbf{23.2_{\color{blue_ppt}{+4.7}}}$ & $\mathbf{46.9_{\color{blue_ppt}{+9.9}}}$ & $\underline{85.4_{\color{blue_ppt}{+23.8}}}$ & $\mathbf{17.3_{\color{blue_ppt}{+3.2}}}$ & $\underline{39.78_{\color{blue_ppt}{+10.36}}}$ & $\underline{25.4_{\color{blue_ppt}{+5.3}}}$ & $23.0_{\color{blue_ppt}{+3.9}}$ & $46.2_{\color{blue_ppt}{+3.5}}$ & $82.8_{\color{blue_ppt}{+20.7}}$ & $17.1_{\color{blue_ppt}{+4.3}}$ & $38.90_{\color{blue_ppt}{+7.54}}$ \\
        
        $x_tx''_v$ & $\mathbf{26.6_{\color{blue_ppt}{+10.7}}}$ & $23.0_{\color{blue_ppt}{+4.5}}$ & $\underline{46.8_{\color{blue_ppt}{+9.8}}}$ & $\mathbf{85.8_{\color{blue_ppt}{+24.2}}}$ & $\underline{17.2_{\color{blue_ppt}{+3.1}}}$ & $\mathbf{39.88_{\color{blue_ppt}{+10.46}}}$ & $\mathbf{26.3_{\color{blue_ppt}{+6.2}}}$ & $\mathbf{24.0_{\color{blue_ppt}{+4.9}}}$ & $\mathbf{48.7_{\color{blue_ppt}{+6.0}}}$ & $\mathbf{90.0_{\color{blue_ppt}{+27.9}}}$ & $\mathbf{17.8_{\color{blue_ppt}{+5.0}}}$ & $\mathbf{41.36_{\color{blue_ppt}{+10.00}}}$ \\
        \hline
    \end{tabular}
    }
    \label{tab:gpt2_cap_all_adapter_large} 
\end{table}

\noindent \textbf{Comparison to cross-attention (complete evaluation)}
We report the full evaluation of the experiments that compare routing functions to cross-attention in the PEFT context in~\cref{tab:gpt_crossattn_all}. 
The shortened version is reported in~\cref{sec:vit_results} (\cref{tab:gpt_crossattn}) in the main text.
The same conclusion as in the main text holds.

\begin{table}[!htbp]
  \centering 
  \caption{Comparing routing functions with cross-attention
  on COCO Cap. $\dag$: parallel Adapter as used in~\cite{ituning2023}. Separate Map.: separate mapping for query/key/value (in cross-attn.) or features $x_R$/$x_H$ (in $x_t(x_v)^\mathsf{T}x_v$). Here $r=128$. Avg. indicates the same average scores of the BLEU-4, METEOR, ROUGE-L, CIDEr and SPICE.} 
  
    \resizebox{\linewidth}{!}{
    \begin{tabular}{lcllllllll}
        \hline
        PEFT & Separate Map. & Alignment & Param. & BLEU-4 & METEOR & ROUGE-L & CIDEr & SPICE & Avg. \\
        \hline
        LoRA & \cmark & Cross-attn. & 4.786M & 28.7 & 24.6 & 52.0 & 92.2 & 17.6 & 43.02 \\
        LoRA & \cmark & $x_t (x_v)^\mathsf{T} x_v$ & \textbf{3.932M} & $30.7_{\color{blue_ppt}{+2.0}}$ & $25.8_{\color{blue_ppt}{+0.8}}$ & $53.4_{\color{blue_ppt}{+1.4}}$ & $99.4_{\color{blue_ppt}{+7.2}}$ & $19.1_{\color{blue_ppt}{+1.5}}$ & $45.68_{\color{blue_ppt}{+2.66}}$ \\
        LoRA & \xmark & $x_t (x_v)^\mathsf{T} x_v$ & \textbf{4.746M} & $30.0_{\color{blue_ppt}{+1.3}}$ & $25.8_{\color{blue_ppt}{+0.8}}$ & $53.1_{\color{blue_ppt}{+1.1}}$ & $99.0_{\color{blue_ppt}{+6.8}}$ & $19.0_{\color{blue_ppt}{+1.4}}$ & $45.38_{\color{blue_ppt}{+2.36}}$ \\

        \hline
         
        $\text{Adapter}^\dag$ & \cmark & Cross-attn. & 4.732M & 30.7 & 25.6 & 53.3 & 99.8 & 19.1 & 45.70 \\
        $\text{Adapter}^\dag$ & \xmark  & $x_t (x_v)^\mathsf{T} x_v$ & \textbf{1.830M} & $30.8_{\color{blue_ppt}{+0.1}}$ & 25.6 & 53.3 & $98.8_{-1.0}$ & $18.9_{-0.2}$ & $45.48_{-0.22}$ \\
        \hline
    \end{tabular}
    }
    \label{tab:gpt_crossattn_all} 
\end{table}

Routing functions in the PEFT modules outperform cross-attention when the number of parameters is similar.
Taking, e.g., $x_t(x_v)^Tx_v$ in Tab. 5, the computation is: $W_{down}x_H(W_{down}x_R)^TW_{down}x_R$; while for cross-attention, we compute as $\phi(W^Q_{down}x_H(W^K_{down}x_R)^T)W^V_{down}x_R$, where $\phi(\cdot)$ refers to a scaling function in cross-attention and $Q$, $K$ and $V$ refer to the query, key and value matrices, respectively. 
As discovered by TextSpan (Gandelsman et al. ICLR2024), the key to maintain the performance of attention lies in the number of dimensions of the embedding space
where they define a lower bound of embedding dimensions to successfully capture the information in the attention heads.
We denote the dimension of the low-rank bottleneck as $r_{W_{down}}$.
Let the dimension of the model be $d$. For cross attention, the number of parameters is $\sum_{i\in\{Q,K,V\}} d \times r_{W^i_{down}} + r_{W^i_{down}}$, 
with the last term referring to the bias term.
For routing functions, the number of parameters is $d \times r_{W_{down}} + r_{W_{down}}$.
Thus, for the same number of parameters, routing functions allow for a larger dimension of the bottleneck (for the example above, roughly by a factor of 3, e.g.  $r_{W^{i}_{down}} \approx r_{W_{down}}/3$). Alternatively, the same bottleneck dimension can be reached with fewer parameters, reducing the risk of overfitting.

\noindent \textbf{Changes in mappings after tuning}
We explore how much does $W_{down}$ and $W_{up}$ change after tuning.
We measure the changes in magnitude of the mapping matrix using the Frobenius norm of the difference as $\Delta W = || W^{t} - W^{0} ||_F$, where $W^{t}$ is the fully trained $W$ and $W^{0}$ is the original $W$. The same computation applies for $W_{up}$ and $W_{down}$.
We further compute $\Delta W_{down} / \Delta W_{up}$ as a measure of which mapping matrix shows more drastic changes after tuning.

\begin{figure}[!htbp]
    \centering
    \begin{minipage}{.5\textwidth}
        \centering
        \includegraphics[width=\linewidth]{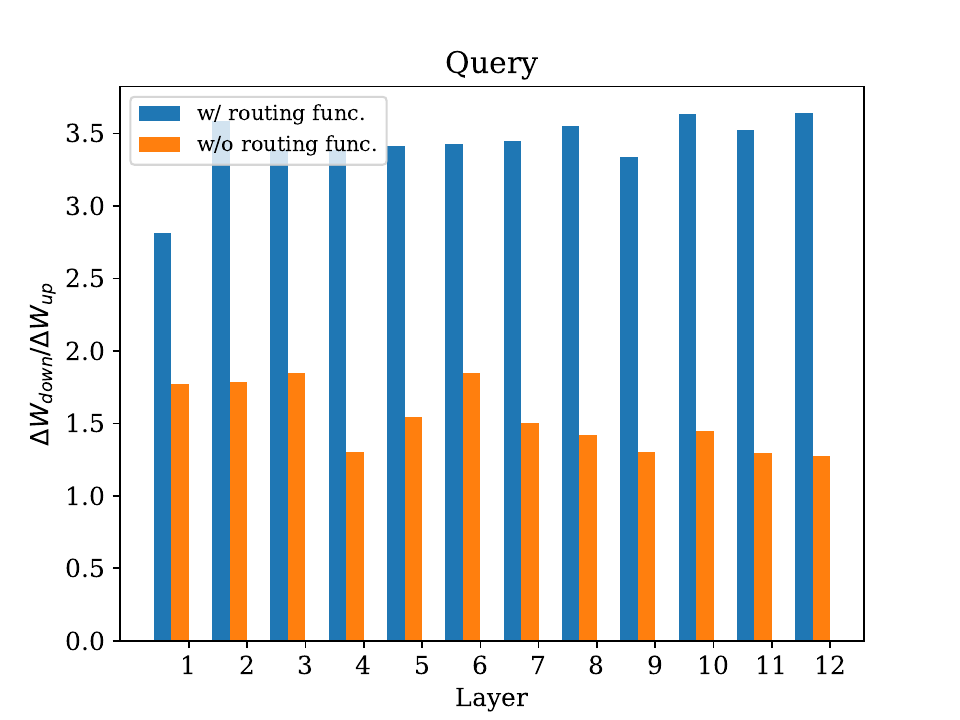}
    \end{minipage}%
    \begin{minipage}{.5\textwidth}
        \centering
        \includegraphics[width=\linewidth]{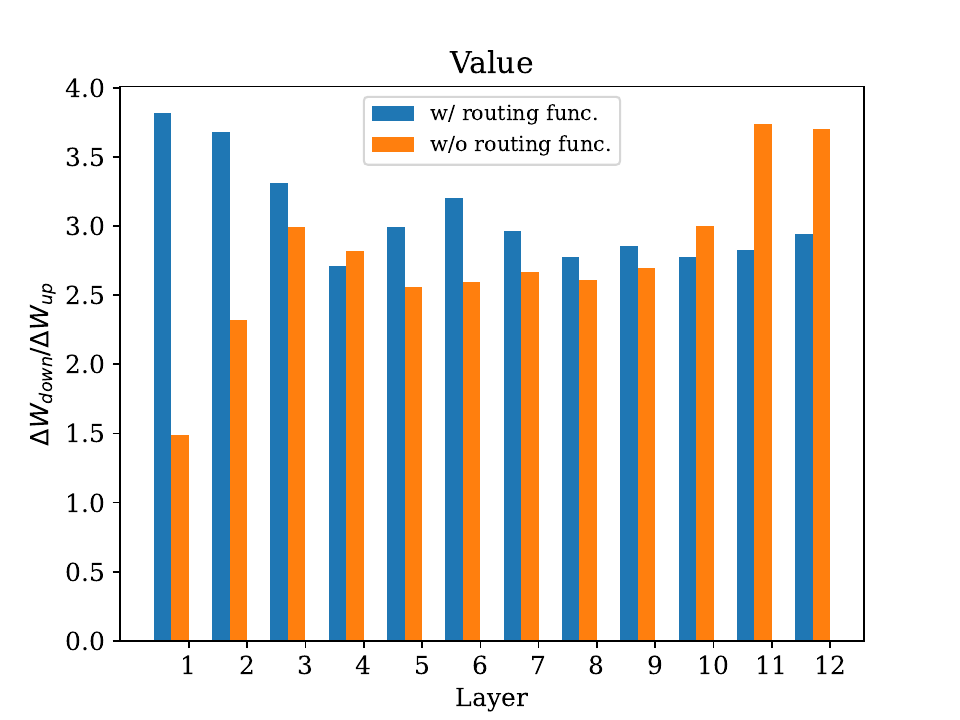}
    \end{minipage}
    \caption{Frobenius norm of the difference ($\Delta W_{down} / \Delta W_{up}$) by layers in GPT2 with or without routing functions. We use the model trained for COCO Cap. with r=64. We use $x_t (x_v)^\mathsf{T}x_v$ with LoRA. The $W_{down}$ and $W_{up}$ mappings are taken from LoRAs added to the the query mapping (Left) and the value mapping (Right) of the attention module in GPT2.}
    \label{fig:fro_norm_gpt}
\end{figure}

As presented in~\cref{fig:fro_norm_gpt}, with the addition of routing function, the changes in down-project mapping $W_{down}$ are more drastic than changes in up-project mapping $W_{up}$.
It shows that the routing functions have bigger impact on $W_{down}$ as training progresses, which is desired as we are learning VL alignment in the low-rank bottleneck.
This intriguing phenomenon reveals properties of the learning dynamics of routing functions.

\section{Additional Results with CLIP-BART}
\label{sec:appendix_results_clip_bart}

In this section we report the standard deviations obtained by routing function ($x_t (x_v)^\mathsf{T}x_v$) and VL-Adapter (as reported in~\cite{Hu_2023_ICCV}). For the detailed evaluation on routing functions with single Adapter/LoRA for CLIP-BART, we refer to~\cref{sec:clip_bart_results} (\cref{tab:clip_bart_single}).


\noindent \textbf{Variations in CLIP-BART experiments}
To verify whether the results produced by the routing functions are robust enough, we are also interested in the variations over different runs.
Following~\cite{Hu_2023_ICCV}, we report the mean and standard deviation over three runs. 
The results from the most performant routing functions are presented in~\cref{tab:clip_bart_significance}.
The improvements in all settings are statistically significant, except for GQA w/ single Adapter. Large gains are seen for $\text{NLVR}^2$ and COCO Cap.

\begin{table}[!htbp]
  \centering 
    \caption{Significance test for best routing functions ($\text{mean}_{\pm\text{std. dev.}}$). We mark the significantly better results in bold. We report the average accuracy of VQA, GQA \& $\text{NLVR}^2$ as Avg. dis. For full results please refer to~\cref{tab:clip_bart_std} and~\cref{tab:clip_bart_multiple_std}.} 
    \resizebox{\linewidth}{!}{
    \begin{tabular}{l l l | llll | l }
        \hline
        & & & VQA & GQA & $\text{NLVR}^2$ & Avg. dis. & COCO Cap. \\
        
        \hline
        \multirow{6}{*}{\rotatebox{90}{\small Single}} & \multirow{3}{*}{\rotatebox{90}{\small LoRA}} & \small None[\textcolor{eccvblue}{11}] &  $65.15_{ \pm0.16}$ & $53.66_{ \pm0.84}$ & $72.58_{ \pm0.73}$ & 63.80 & $115.01_{ \pm0.26}$ \\

        \cline{3-8}
        
        & & routing &  $\mathbf{65.68_{   \pm0.01}}$  & $\mathbf{53.96_{ \pm0.12}}$  & $\mathbf{73.60_{ \pm0.36}}$  & $\mathbf{64.35_{ \pm0.22}}$  & $\mathbf{119.26_{ \pm0.25}}$  \\

        & & function &  \small ($x_t\circ x'_v$) &  \small ($x_t\circ x'_v$) &  \small ($x_t(x_v)^\mathsf{T}x_v$) & \small ($x_t\circ x'_v$) & \small ($x_tx''_v$) \\

        \cline{2-8}

        & \multirow{3}{*}{\rotatebox{90}{ \small Adpt.}} & \small None[\textcolor{eccvblue}{11}] &  $65.76_{\pm0.28}$ & $54.16_{\pm0.44}$  & $73.19_{\pm0.71}$ & 64.37 & $114.61_{ \pm0.26}$ \\

        \cline{3-8}
        
        & & routing &  $\mathbf{65.92_{\pm0.20}}$  & $54.54_{\pm0.24}$ & $\mathbf{74.52_{\pm0.36}}$  & $\mathbf{64.83_{\pm0.05}}$  & $\mathbf{118.86_{\pm0.15}}$ \\

        & & function & \small ($x_t \circ x'_v$) & \small ($x_t \circ x'_v$) &  \small ($x_t \circ x'_v$) & \small ($x_t \circ x'_v$) &  \small ($x_t x''_v$) \\

        \hline
        \multirow{3}{*}{\rotatebox{90}{\small Multi.}} & \multirow{3}{*}{\rotatebox{90}{\small Adpt.}} & \small None[\textcolor{eccvblue}{37}] & 65.4 & 54.0 & 69.8 & 63.0 & 114.3  \\
        
        \cline{3-8}
        & & routing & $\mathbf{66.39_{\pm0.04}}$ & $\mathbf{54.22_{\pm0.19}}$ & $\mathbf{71.83_{\pm0.80}}$ & $\mathbf{63.96_{\pm0.26}}$ & $\mathbf{117.81_{\pm0.75}}$  \\

        & & function & \small (comb1) & \small (comb1) & \small (comb3) & \small (comb3) & \small (comb1) \\

        \hline
    \end{tabular}
    }
    \label{tab:clip_bart_significance} 
\end{table}

\begin{table}[!htbp]
  \centering 
    \caption{Experiments with $x_v(x_v)^\mathsf{T}x_v$ on multitask learning (VQA, GQA, $\text{NLVR}^2$ and COCO Cap.) using CLIP-BART with single LoRA/Adapter. We report the CIDEr score for COCO Cap. and accuracy for the other three tasks. We take the baseline results from \cite{Hu_2023_ICCV} ($\dag$). We report the average accuracy of VQA, GQA \& $\text{NLVR}^2$ as Avg. dis. and the average score of all four tasks as Avg. We report the mean and standard deviation over three runs.} 
    \resizebox{\linewidth}{!}{
    \begin{tabular}{lllllll | llllll}
        \hline
        Routing & \multicolumn{5}{c}{Single LoRA} & \multicolumn{5}{c}{Single Adapter} \\
        Functions & VQA & GQA & $\text{NLVR}^2$ & COCO Cap. & Avg. & Avg. dis. & VQA & GQA & $\text{NLVR}^2$ & COCO Cap. & Avg. & Avg. dis. \\
        \hline
        $\text{None}^\dag$ & $65.15_{\pm0.16}$ & $53.66_{\pm0.84}$ & $72.58_{\pm0.73}$ & $115.01_{\pm0.26}$ & $76.60_{\pm0.32}$ & 63.80 & $65.76_{\pm0.28}$ & $54.16_{\pm0.44}$  & $73.19_{\pm0.71}$ & $114.61_{\pm0.26}$ & $76.93_{\pm0.25}$ & 64.37 \\

        $x_t \circ x'_v$ & $65.68_{\pm0.01}$ & $53.96_{\pm0.12}$ & $73.42_{\pm0.63}$ & $113.94_{\pm0.43}$ & $76.75_{\pm0.06}$ & $64.35_{\pm0.22}$ & $65.92_{\pm0.20}$ & $54.34_{\pm0.17}$ & $74.23_{\pm0.23}$ & $114.38_{\pm0.60}$ & $77.22_{\pm0.16}$ & $64.83_{\pm0.05}$ \\

        $x_t + x'_v$ & $65.14_{\pm0.10}$ & $53.73_{\pm0.26}$ & $73.51_{\pm0.22}$ & $114.96_{\pm0.53}$ & $76.84_{\pm0.07}$ & $64.13_{\pm0.13}$ & $65.89_{\pm0.10}$ & $54.18_{\pm0.53}$ & $73.90_{\pm0.28}$ & $114.39_{\pm0.77}$ & $77.09_{\pm0.35}$ & $64.66_{\pm0.21}$ \\

        $x_t(x_v)^\mathsf{T}x_v$ & $64.94_{\pm0.09}$ & $53.56_{\pm0.31}$ & $73.60_{\pm0.36}$ & $117.80_{\pm0.34}$ & $77.48_{\pm0.11}$ & $64.03_{\pm0.14}$ & $65.84_{\pm0.32}$ & $53.65_{\pm0.47}$ & $74.31_{\pm0.09}$ & $117.65_{\pm0.39}$ & $77.87_{\pm0.18}$ & $64.03_{\pm0.14}$ \\

        $x_tx''_v$ & $64.84_{\pm0.06}$ & $53.13_{\pm0.37}$ & $72.98_{\pm0.44}$ & $119.26_{\pm0.25}$ & $77.55_{\pm0.10}$ & $63.65_{\pm0.06}$ & $65.83_{\pm0.22}$ & $53.61_{\pm0.33}$ &  $73.27_{\pm1.24}$ & $118.86_{\pm0.15}$ & $77.90_{\pm0.25}$ & $64.24_{\pm0.38}$ \\
        
        \hline
    \end{tabular}
    }
    \label{tab:clip_bart_std} 
\end{table}

\begin{table}[!htbp]
  \centering 
    \caption{Experiments with different combinations of routing functions on multitask learning using CLIP-BART with multiple Adapters. 
    We report CIDEr score for COCO Cap., accuracy for the other three tasks.
    We take the baseline results from \cite{Sung_2022_CVPR} ($\ddag$). We report the average accuracy of VQA, GQA \& $\text{NLVR}^2$ as Avg. dis. and the average score of all four tasks as Avg. We report the mean and standard deviation over three runs.} 
    \resizebox{\linewidth}{!}{
    \begin{tabular}{l | c | c | c | c | llllll}
        \hline
        Routing & \multicolumn{4}{c}{Routing Functions} & \multicolumn{5}{c}{Performance} \\
        Functions & VQA & GQA & $\text{NLVR}^2$ & COCO Cap.  & VQA & GQA & $\text{NLVR}^2$ & COCO Cap. & Avg. & Avg. dis.\\
        \hline
        $\text{None}^\ddag$ & \xmark & \xmark & \xmark & \xmark & 65.4 & 54.0 & 69.8 & 114.3 & 75.9 & 63.07 \\
        
        \hline

        comb1 & $x_t\circ x'_v$ & $x_t\circ x'_v$ & $x_t(x_v)^\mathsf{T}x_v$ & $x_tx''_v$ & $66.39_{\pm0.04}$ & $54.11_{\pm0.37}$ & $69.97_{\pm0.74}$ & $117.81_{\pm0.75}$ & $77.07_{\pm0.37}$ & $63.49_{\pm0.31}$ \\

        comb2 & $x_t+x'_v$ & $x_t+x'_v$ & $x_t(x_v)^Tx_v$ & $x_tx''_v$ & $65.04_{\pm0.20}$ & $53.00_{\pm0.40}$ & $70.96_{\pm0.41}$ & $117.00_{\pm0.51}$ & $76.50_{\pm0.21}$ & $63.07_{\pm0.11}$ \\

        comb3 & $x_t\circ x'_v$ & $x_t\circ x'_v$ & $x_t\circ x'_v$ & $x_tx''_v$ & $66.28_{\pm0.20}$ & $53.77_{\pm0.45}$ & $71.83_{\pm0.80}$  & $117.66_{\pm0.67}$ & $77.46_{\pm0.35}$ & $63.96_{\pm0.26}$ \\

        comb4 & $x_t + x'_v$ & $x_t + x'_v$ & $x_t\circ x'_v$ & $x_tx''_v$ & $65.32_{\pm0.13}$ & $53.15_{\pm0.15}$ & $71.00_{\pm0.11}$ & $117.40_{\pm0.93}$ & $76.72_{\pm0.21}$ & $63.16_{\pm0.07}$ \\

        \hline
    \end{tabular}
    }
    \label{tab:clip_bart_multiple_std} 
\end{table}

As presented in~\cref{tab:clip_bart_std} and~\cref{tab:clip_bart_multiple_std}, the results obtained by the routing function are consistent.
We even obtain much lower standard deviations in many tasks (\eg on GQA\&$\text{NLVR}^2$ for Single LoRA, on $\text{NLVR}^2$ for Single Adapter). 
We also obtain lower standard deviations on average scores for both Single LoRA and Single Adapter.
The results show that our method is robust with low variations in performance.
